\newcommand{\GSF}{\tilde{S}}

\documentclass[10pt,twocolumn,letterpaper]{article}

\usepackage{cvpr}      
\definecolor{cvprblue}{rgb}{0.21,0.49,0.74}
\usepackage[pagebackref,breaklinks,colorlinks,allcolors=cvprblue]{hyperref}
\usepackage{caption} 
\usepackage{mathtools} 
\usepackage{bbold} 
\usepackage[accsupp]{axessibility} 

\usepackage{booktabs}    
\usepackage{colortbl}    
\usepackage{xcolor}      
\usepackage{caption}     

\def\paperID{***} 
\def\confName{CVPR}
\def\confYear{2026}

\title{SGSoft: Learning Fused Semantic-Geometric Features for 3D Shape Correspondence via Template-Guided Soft Signals}

\author{
Soyeon Yoon$^{1}$ \quad
Chang Wook Seo$^{2}$ \quad
Hyunjung Shim$^{1}$\\[0.5em]
$^{1}$KAIST AI \\
$^{2}$Anigma Technologies \\[0.5em]
{\tt\small thoyeony@kaist.ac.kr \quad lgtwins@anigma-ai.com \quad kateshim@kaist.ac.kr}
}

\begin{document}
\twocolumn[{
\maketitle
\begin{center}
    \vspace{-7mm}
    \makebox[\textwidth][c]{%
        \includegraphics[width=1.05\textwidth]{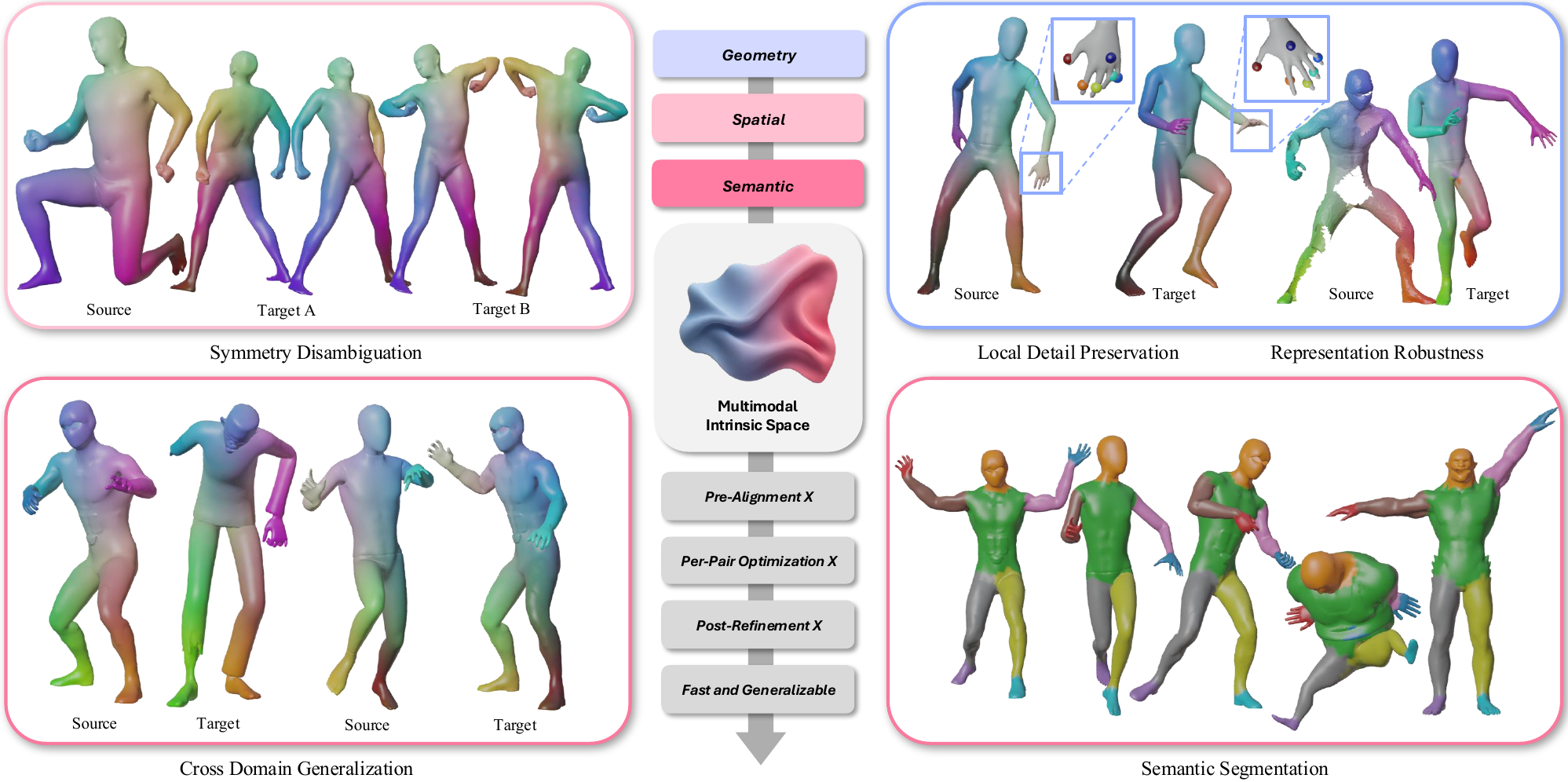}
    }
    \vspace{-5mm} 
    \captionof{figure}{
    \textbf{Visualization of geometric and semantic fidelity of SGSoft under variations in pose, topology, domain, and representation.}
    The framework operates without pre-alignment, pairwise optimization or post-refinement, enabling fast and generalizable correspondence retrieval.
    This illustrates (top-left) symmetry disambiguation, (bottom-left) cross-domain generalization, (top-right) local detail preservation and representation robustness, and (bottom-right) descriptor versatility for semantic segmentation.
    }
    \label{fig:teaser}
\end{center}

}]
\begin{abstract}
Learning dense correspondences across deformable 3D shapes remains a long-standing challenge due to structural variability, non-isometric deformation, and inconsistent topology. Existing methods typically trade off generalization, geometric fidelity, and efficiency.
We address this by proposing SGSoft, a unified intrinsic pipeline that (i) constructs a geodesic correspondence field on a canonical template, (ii) learns multimodal dense descriptors guided by pretrained semantic priors with this geodesic correspondence field supervision, (iii) retrieves dense correspondences in a single feed-forward pass via nearest-neighbor search in descriptor space.
This formulation enables stable and topology-invariant supervision under large pose variation, structural differences, and remeshing.
SGSoft achieves state-of-the-art inter-category generalization while offering the best accuracy–efficiency trade-off among prior methods. It also achieves near real-time inference without pre-alignment, pairwise optimization, or post-refinement. 
Learned descriptors can be transferred effectively to downstream tasks such as semantic segmentation and deformation transfer, establishing a scalable and deployment-ready paradigm for dense 3D correspondence. 
\end{abstract}
    
\vspace{-3mm}
\section{Introduction}
\label{sec:intro}

Dense correspondence between 3D shapes aims to establish point-to-point mappings that preserve geometric fidelity and semantic alignment under variation in pose, topology, and fine-scale structure. This capability is fundamental to a wide range of downstream applications, including motion retargeting~\cite{ye2024skinned,sun2022human}, rig~\cite{xu2020rignet,paravati2016point} and texture~\cite{mertens2006texture} transfer, shape interpolation~\cite{deng2021deformed,zheng2021deep} and more. 
However, real-world pipelines routinely involve unaligned scans, remeshed surfaces, and domain shifts, which render manual registration and post-refinement prohibitively costly and inherently non-scalable. 
This motivates the need for an automated correspondence framework that is accurate, robust, and computationally efficient, with the capacity to operate over large shape repositories and support real-time interaction.

Achieving such a system requires resolving two core challenges: generalization and efficiency.
Generalization refers to the ability to produce consistent correspondences under changes in alignment, pose, scale, topology, or domain. Existing approaches struggle under this variability and therefore rely on pre-alignment, pairwise optimization, or post-refinement as compensatory steps. However, these mechanisms do not address the underlying generalization issue and often fail under large deformations or cross-domain shifts.

Efficiency, on the other hand, concerns inference speed, pipeline complexity, and dependence on refinement. Prior methods rely on manual pre-alignment or pairwise optimization to compensate for limited generalization, which causes inference cost to grow rapidly and become impractical at scale. Consequently, approaches requiring pre-alignment~\cite{liu2025stable}, pairwise optimization~\cite{pierson2025diffumatch}, or multi-view rendering~\cite{dutt2024diffusion}, or post-refinement~\cite{zhuravlev2025denoising} lose practical viability in real-world settings.

Prior work in dense correspondence falls into four categories. \emph{Deformation-based methods}~\cite{aigerman2022neural,amberg2007optimal,sorkine2007rigid,groueix20183d,eisenberger2019divergence,ehm2024geometrically,yang2023gencorres} optimize explicit deformations and achieve accurate local alignment, but require careful initialization and per-pair optimization, limiting scalability and generalization. \emph{Functional map approaches}~\cite{ovsjanikov2012functional,litany2017deep,sun2023spatially,yona2025neural,donati2020deep} provide compact, globally consistent mappings in the spectral domain, yet lose fine geometric/semantic detail and remain sensitive to noise and topology, often necessitating refinement or candidate selection.
\emph{Large 2D vision model–based methods}~\cite{chen2025dv,uzolas2025surface,zhu2024densematcher,zhang2024telling,dutt2024diffusion} inherit strong semantic priors from large-scale pretraining but suffer from occlusion and depth ambiguities introduced by multi-view rendering and back-projection, leading to degraded geometric fidelity and high computational overhead.
\emph{Hybrid methods}~\cite{eisenberger2020smooth,eisenberger2020deep,attaiki2022ncp,sun2023spatially,pierson2025diffumatch,zhuravlev2025denoising,liu2025stable,rimon2025fridu} combine multiple paradigms to mitigate individual weaknesses, yet still rely on alignment, post-refinement, or heavy computation, preventing a truly scalable and efficient solution.

To address these limitations, we propose a novel unified framework that addresses both the generalization and efficiency challenges of dense shape correspondence. Unlike deformation-based registration or spectral functional-map pipelines, our method constructs a multimodal intrinsic correspondence space grounded in a template-based geodesic correspondence field.
Our approach defines this geodesic correspondence field as a supervisory signal and uses it to learn a multimodal dense descriptor that fuses (i) geometric, (ii) semantic, and (iii) spatial awareness in a single representation.
The geodesic correspondence field, as (i) a geometric component, is a geodesic-informed probabilistic field defined on a canonical template. Unlike hard anchors, which are sensitive to topology changes and remeshing during training, this field maintains continuous, topology-invariant correspondences. This stability enables flexible and intrinsic mapping of diverse inputs into the canonical space. 

However, the geodesic correspondence field alone does not encode fine-grained semantic cues and high-level spatial understanding needed to disambiguate left–right or front–back correspondences. To bridge this gap, we initialize our descriptor with (ii) semantic priors from a pretrained 3D vision foundation model (e.g., Uni3D~\cite{zhou2023uni3d}), then further finetune it using three objectives. These include (i) a  contrastive loss guided by the geodesic correspondence field to maintain local and global geometric coherence, (ii) a part classification loss that reinforces coarse semantic segmentation and stabilizes initial training, and (iii) a symmetry-aware loss that suppresses correspondence leakage across left–right or front–back symmetric regions. Through this combination of supervisory signals, our framework learns a unified dense descriptor capable of high semantic precision, symmetry disambiguation, and geometric robustness across diverse shapes, poses, and domains.

At inference, SGSoft computes dense descriptors in a single forward pass, requiring no pre-alignment, optimization, post-refinement, rendering, or candidate selection. 
This lightweight pipeline achieves near real-time performance while preserving strong inter-category generalization, resolving both efficiency and generalization challenges without relying on compensatory alignment or optimization steps.

In summary, our contributions are threefold: 
(1) introducing a geodesic correspondence field that provides stable, continuous, and topology-invariant supervision for dense correspondence learning (2) proposing a multimodal dense descriptor that combines strong cross-category generalization with geometric and spatial awareness learned through the geodesic correspondence field and its training recipe, and (3) integrating the entire process into a fast, alignment-free, optimization-free, and post-refinement-free correspondence framework that is suitable for practical deployment while preserving robust generalization.

\vspace{-3mm}
\section{Related Work}
\label{sec:related_work}

\subsection{3D Shape Correspondence Methods}
\label{sec:related_matching}
Establishing dense 3D correspondence has long been a fundamental yet challenging problem, with research evolving across several major paradigms. 
\emph{Deformation-based methods} spatially align shapes through explicit deformation, either via iterative optimization (e.g., ICP~\cite{amberg2007optimal}, ARAP~\cite{sorkine2007rigid}) or by directly learning deformation fields through neural networks~\cite{groueix20183d,huang2021arapreg,ehm2024geometrically,bernard2020mina,aigerman2022neural, marin2024nicp}. 
Recent variants incorporate generative priors to regularize the deformation process~\cite{yang2023gencorres}, achieving high accuracy but remaining limited to per-pair alignment and sensitive to initialization. 

\emph{Functional map (FM)-based methods}~\cite{ovsjanikov2012functional,cao2022unsupervised,bracha2024wormhole,bracha2024unsupervised,attaiki2023shape,donati2020deep,donati2022complex,ehm2024partial, cao2023unsupervised} instead represent correspondences as linear operators between function spaces, allowing compact and globally consistent reasoning.
Learning-based methods that learn robust descriptors directly from geometry~\cite{donati2020deep} or use generative models to predict the functional map itself~\cite{zhuravlev2025denoising} enhance spectral robustness and continuity. However, their spectral abstraction inherently suppresses fine-grained geometric or semantic detail, making them difficult to distinguish symmetric or closely related parts. 

\emph{Large 2D vision model-based methods} mark a recent shift toward semantic-aware correspondence by leveraging large-scale pre-trained models~\cite{darcet2023vitneedreg,oquab2023dinov2,caron2021emerging,podell2023sdxl}. Recent works~\cite{dutt2024diffusion, zhu2024densematcher} extract multi-view image features and unproject them into 3D, yielding strong inter-category generalization and high-level semantic understanding. 
However, their dependency on rendered supervision makes them sensitive to occlusion, view sparsity, and texture-free surfaces, while the heavy computational cost of diffusion-based priors limits scalability and prevents real-time deployment.

Recent efforts have begun exploring \emph{hybrid systems}~\cite{attaiki2022ncp,bastian2024hybrid,cao2024spectral,cao2024synchronous,eisenberger2021neuromorph,eisenberger2020smooth,sun2023spatially,liu2025stable,rimon2025fridu} that integrate complementary strengths across the aforementioned paradigms.
While not a formally established category, we use the term hybrid to describe approaches that explicitly couple the geometric fidelity of FM or deformation-based models with the semantic priors of large-scale 2D vision backbones.
Such designs aim to bridge low-level geometric alignment and high-level semantic understanding within a single framework, yet often sacrifice efficiency and architectural coherence in the process.
Despite these advances, existing paradigms still treat geometric alignment, spatial reasoning, and semantic awareness as largely separate objectives.
We introduce SGSoft, a multimodal intrinsic correspondence framework that unifies geodesic geometry, semantic priors, and spatial cues within a single feed-forward model.

\subsection{Template-Guided Supervision}
\label{sec:related_template}




Leveraging a canonical template as a shared reference space has become a central strategy for enforcing consistent correspondence across diverse shapes and poses. Parametric human models such as SMPL~\cite{Loper2015} provide dense pseudo-ground-truth labels that have enabled a wide range of supervised tasks~\cite{saito2019pifu,xiu2022icon}. In parallel, neural field methods~\cite{zheng2021deep,deng2021deformed,liu2023unsupervised,attaiki2022ncp,kim2023semantic,liu2024self,liu2023learning,zhang2023self} learn canonical latent spaces by warping individual geometries into a shared volumetric or feature domain, implicitly encoding correspondences through deformation fields. 

Building on these complementary directions, we introduce a multimodal intrinsic correspondence space that retains the canonical template as a stable intrinsic reference while integrating geometric, semantic, and spatial cues from neural representations. This intrinsic space provides consistent surface-level supervision and preserves fine geometric detail, yet flexibly accommodates large shape and pose variation within a unified feed-forward framework.

\vspace{-1mm}
\subsection{3D Foundation Models for Semantic Understanding}
\label{sec:related_feature}

Semantic understanding from 3D input is a core objective in 3D vision, driving progress from local geometric processing to large-scale representation learning. Early point-based and Transformer-based networks~\cite{qi2017pointnet,qi2017pointnet++,yu2022point,guo2021pct} capture global structure and contextual relationships, while recent multimodal 3D foundation models align 3D features with image–language representations~\cite{xue2023ulip,guo2023point,zhou2023uni3d,qi2024unigs,liu2024uni3d}, enabling richer semantic priors. Among them, Uni3D~\cite{zhou2023uni3d} provides a scalable unified framework that bridges 2D vision and 3D geometry, producing semantically meaningful and transferable representations.

However, these foundation features often lack the spatial precision and geometric consistency required for dense correspondence. To address this, our framework adopts Uni3D as a semantic backbone and, to our knowledge, is the first to unify complementary semantic, geometric, and spatial cues into a single descriptor. By refining its features with geometry-aware and spatially discriminative objectives, we produce a descriptor that preserves strong semantic expressiveness while gaining the intrinsic geometric structure and spatial disambiguation essential for accurate correspondence
\section{Methodology}
\label{sec:method}
\subsection{Overview}
\label{sec:method:overview}
Our framework, SGSoft, establishes generalizable and efficient 3D correspondence through four synergistic components. 
First, the geodesic correspondence field $\GSF$ provides stable, topology-invariant geometric supervision that preserves surface continuity and smoothness across discretizations (Sec.~\ref{sec:method:soft}). 
Second, the multimodal descriptor module fuses complementary signals, including semantic priors from a pretrained 3D foundation model, Uni3D~\cite{zhou2023uni3d}, intrinsic geometric cues from $\GSF$, and spatial disambiguation from symmetry-aware supervision, within a shared intrinsic template space (Sec.~\ref{sec:method:descriptor}).
Third, training objectives including geodesic weighted contrastive loss, part classification loss, and symmetry loss reinforce geometric coherence and semantic distinctiveness (Sec.~\ref{sec:method:loss}). 
Finally, a lightweight inference pipeline predicts dense correspondences in a single forward pass without any pre-alignment optimization or post refinement (Sec.~\ref{sec:method:train_infer}). 
Together, these components define a multimodal intrinsic correspondence space that is semantically discriminative, geometrically robust, and stable under large non rigid deformations.

\begin{figure*}
    \vspace{-6mm}
    \makebox[\textwidth][c]{%
        \includegraphics[width=1.05\textwidth]{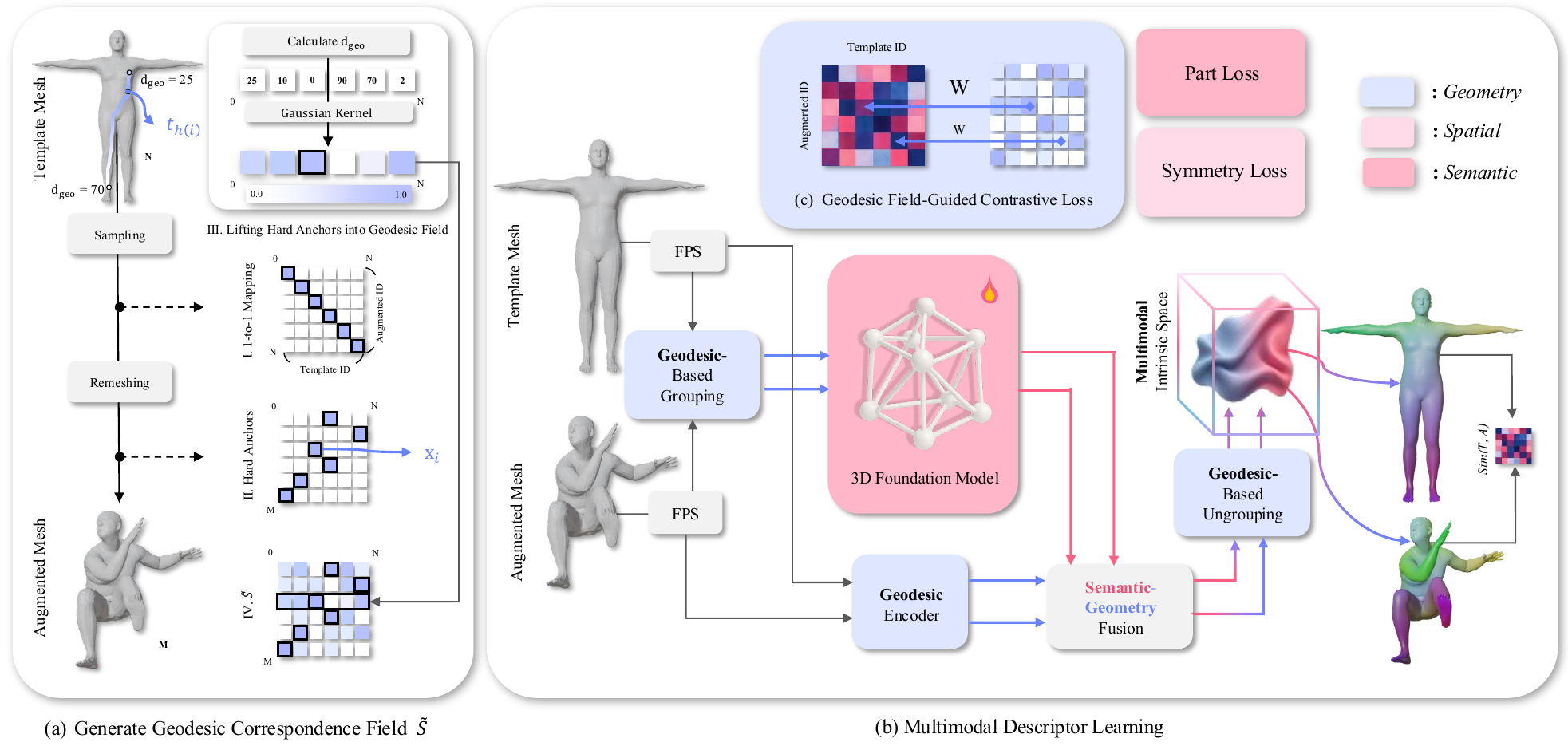}
    }
    \vspace{-5mm}
    \captionof{figure}{
    \textbf{Overview of SGSoft.}
    (a) A geodesic correspondence field  \(\GSF\)\ provides topology- and deformation-robust supervision by propagating correspondence likelihoods from anchor vertices across the canonical template surface. 
    (b) Multimodal descriptors are trained under \(\GSF\)\ supervision by fusing semantic, geometric, and spatial cues through local geodesic grouping and ungrouping, global geodesic encoding, transformer-based fusion, and training objectives. 
    (c) A geodesic correspondence field-guided contrastive loss aligns per-vertex descriptors, weighting correspondence likelihoods by geodesic proximity to enforce a geodesically discriminative embedding space.
    }
    \vspace{-5mm}
    \label{fig:main}
\end{figure*}

\subsection{Geodesic Correspondence Field}
\label{sec:method:soft}
\paragraph{Motivation}
Learning dense correspondences requires stability under remeshing and large non-isometric deformations while maintaining local surface continuity.
We address this using two intrinsic properties of surface geometry:
\textbf{(I1)} Geodesic distances are deformation-invariant and capture intrinsic relationships between surface points.
\textbf{(I2)} Geodesic probabilistic fields defined on a template preserve local smoothness, remain robust to mesh perturbations, and provide a consistent mapping into a canonical intrinsic space.
We therefore reformulate correspondence learning as aligning geodesic probabilistic fields within this intrinsic canonical space, providing stable supervision across diverse shapes, poses, and discretizations.

\vspace{-2mm}
\paragraph{Hard Anchors Across Diverse Template Variants.}
We first define vertex-level anchor pairs between a canonical template and its parametric variants~\cite{Loper2015}. 
To ensure robustness against topology and indexing changes, we apply augmentations (subdivision, decimation, random rotation) and recover hard anchors by back-tracing the augmentation chain. 
Each augmented vertex \(x_i\) maps to a canonical vertex \(t_{h(i)}\), producing consistent anchor sets and corresponding part labels \(\ell_\text{aug}\) across all remeshed or deformed variants.

\vspace{-2mm}
\paragraph{Lifting Hard Anchors into Geodesic Fields.}
Each hard anchor is lifted into a continuous geodesic field on the canonical template:
\vspace{-2mm}
\begin{equation}
\tilde{S}_{i,v} = \exp\!\left(-\frac{d_\mathrm{geo}(t_{h(i)}, t_v)^2}{\sigma^2}\right),
\end{equation}
where \(d_\mathrm{geo}(\cdot,\cdot)\) is the geodesic distance and \(\sigma\) controls locality. 
This gaussian kernel defines a smooth intrinsic field centered at \(t_{h(i)}\), assigning higher weights to vertices \(t_{v}\) that are geodesically close and exponentially attenuating those farther away. Since the field is defined in template space, it provides continuous, topology-invariant supervision across deformations.
As geometric supervision, each augmented mesh point references its mapped template location and takes its geodesic correspondence field value as intrinsic guidance (Fig.~\ref{fig:main} (a)).

We further apply adaptive modulation that adjusts field coverage according to vertex density and semantic part consistency, yielding a sparser and more efficient soft-field map.  
Additionally, local curvature is used to weight vertex contributions, emphasizing structurally salient regions such as joints and facial features (see Supp.~A.1).

\subsection{Multimodal Descriptor Learning under Geodesic Supervision}
\label{sec:method:descriptor}
With \(\GSF\)\ as supervision, we train multimodal descriptors that integrate semantic, geometric, and spatial information (Fig.~\ref{fig:main} (b)).
To enhance generalization and semantic grounding, we initialize our model with the pretrained 3D foundation model Uni3D~\cite{zhou2023uni3d}, which encodes semantic understanding directly from 3D point clouds. 
The rationale for selecting Uni3D as the backbone is detailed in Supp.~G.2.
All components described in this section are applied consistently to both the canonical template and all augmented training meshes.

\vspace{-3mm}
\paragraph{Geodesic Grouping and Propagation.}
We partition the mesh into $N$ geodesic patches using furthest-point sampling (FPS)~\cite{qi2017pointnet++}. 
For each patch center $c_g$, we assign the $M$ nearest vertices 
based on geodesic distance:
\begin{equation}
\mathcal{N}(c_g) = 
\operatorname{TopM}_{x_i}\!\big(-d_\mathrm{geo}(x_i, c_g)\big),
\label{eq:geodesic_group}
\end{equation}
where $d_\mathrm{geo}(x_i, c_g)$ denotes the intrinsic geodesic distance 
between vertex $x_i$ and patch center $c_g$, and $M$ is a fixed neighborhood 
size that controls the number of vertices in each patch.

These geodesic groupings later serve as the structural basis for propagating fused patch embeddings \(z_g\), which combine semantic features with intrinsic geometry. The fused embeddings are transferred to each vertex via 
geodesic-weighted interpolation, where each vertex aggregates 
features from its $k$ nearest patch centers in geodesic space:
\begin{equation}
z(x_i) =
\frac{\sum_{g \in \mathcal{N}_k(i)} w_{ig}\, z_g}
     {\sum_{g \in \mathcal{N}_k(i)} w_{ig}},
\quad
w_{ig} = \frac{1}{d_\mathrm{geo}(x_i, c_g)^p + \epsilon}.
\label{eq:geodesic_ungroup}
\end{equation}
This transfers local geometric context to vertices while preserving intrinsic surface continuity and preventing feature leakage into adjacent parts.

\vspace{-3mm}
\paragraph{Geodesic Feature Encoding.}
To capture global intrinsic structure beyond local grouping, we encode pairwise geodesic relations among patch centers into embeddings.  
Given the set of patch centers \(C = \{c_1, c_2, \dots, c_G\}\), we compute a geodesic distance matrix:
\begin{equation}
D_\mathrm{geo}(i,j) = d_\mathrm{geo}(c_i, c_j),
\label{eq:geodesic_matrix}
\end{equation}
where \(d_\mathrm{geo}(\cdot,\cdot)\) denotes the intrinsic geodesic distance between patch centers \(c_i\) and \(c_j\).  
We then vectorize the upper triangular part of \(D_\mathrm{geo}\) to obtain a compact representation of pairwise geodesic connectivity:
\begin{equation}
v_\mathrm{geo} = [\,d_\mathrm{geo}(c_i, c_j) \mid i < j\,].
\label{eq:geodesic_vector}
\end{equation}
This vector is projected through a transformer~\cite{vaswani2017attention} encoder that models higher-order dependencies among patch centers and produces a global geodesic embedding:
\begin{equation}
g_\mathrm{geo} = \mathrm{Encoder}(v_\mathrm{geo}).
\label{eq:geodesic_encoder}
\end{equation}
The resulting feature \(g_\mathrm{geo}\) serves as a global intrinsic geometry context for robust correspondence learning.

\paragraph{Semantic Prior Fusion with Geodesic Refinement.}
\label{sec:method:fusion}
Finally, we unify Uni3D semantic representations with intrinsic geometric structure through cross-modal fusion. 
The patch-level semantic embedding $f_{\mathrm{sem}}(c_g)$ extracted from Uni3D is combined with the global geodesic feature $g_{\mathrm{geo}}$ (Eq.~\ref{eq:geodesic_encoder}) to produce fused patch descriptors:
\begin{equation}
z_g = \Gamma\!\big(\,[\,f_{\mathrm{sem}}(c_g)\,\|\,g_{\mathrm{geo}}\,]\,\big),
\label{eq:fusion_descriptor_patch}
\end{equation}
where $\Gamma(\cdot)$ denotes a fusion module that adaptively balances semantic and geometric contributions. 
The resulting patch descriptors $\{z_g\}$ are then propagated to individual vertices through geodesic-weighted ungrouping (Eq.~\ref{eq:geodesic_ungroup}), yielding the final per-vertex descriptors $z(x)$. 
This formulation aligns high-level semantic cues with both local and global intrinsic geometry, producing descriptors that are semantically discriminative and geometrically consistent across diverse poses, resolutions, and deformations.

\subsection{Training Objectives}
\label{sec:method:loss}
We supervise the fused semantic and geometric point descriptors \(z(x)\) using three objectives that enforce geodesic locality across deformations and discretizations, semantic discriminability, and symmetry-aware spatial consistency.
\vspace{-2mm}
\paragraph{Geodesic Correspondence Field–Guided Contrastive Loss.}  
The geodesic correspondence field provides a topology-consistent supervision by distributing correspondence likelihoods over intrinsic neighborhoods on the template surface, encouraging smooth and deformation-robust alignment.  
To this end, we propose a \textbf{geodesic correspondence field-weighted InfoNCE} loss, where the \(\GSF\)\ guides descriptor alignment by weighting correspondences according to their geodesic proximity:
\begin{equation}
\mathcal{L}_\text{soft}
= -\frac{1}{M}\sum_{i,v} \tilde{S}_{i,v}
\log \frac{\exp(\mathbf{A}_{i,v}/\tau)}{\sum_u \exp(\mathbf{A}_{i,u}/\tau)},
\end{equation}
where \(\mathbf{A}\) denotes the cosine similarity between augmented and template descriptors, and \(M\) is the number of vertices on the augmented mesh.  
Unlike standard InfoNCE~\cite{oord2018representation}, our formulation weights correspondences by their geodesic proximity, emphasizing geometrically consistent matches.
This geodesic weighting smooths the embedding space along the intrinsic surface manifold, leading to locally coherent correspondences under diverse topologies and deformations (Fig.~\ref{fig:main} (c)).  

\leavevmode\vspace{-7mm}
\paragraph{Part and Symmetry Loss.}
To enhance semantic part coherence and resolve symmetry-induced ambiguities, we incorporate two auxiliary objectives.  
A part classification loss provides coarse semantic alignment,  enforcing consistent part labels between the augmented and template shapes:
\begin{equation}
\mathcal{L}_\text{part}
= \tfrac{1}{2}\big(
\mathrm{CE}(\mathbf{P}_\text{aug}, \ell_\text{aug})
+
\mathrm{CE}(\mathbf{P}_\text{tmp}, \ell_\text{tmp})
\big),
\end{equation}
where \(\mathrm{CE}(\cdot)\) denotes the cross-entropy loss with label smoothing, 
\(\mathbf{P}\) represents the predicted part segmentation logits, 
and \(\ell\) denotes the corresponding ground-truth part labels for the augmented and template meshes.

We introduce a symmetry regularizer that encourages the descriptor space to distinguish bilaterally symmetric regions (e.g., left and right limbs) while preserving spatial coherence across the surface.
Let \(\mathbf{A}\) be the cosine similarity matrix between augmented and template descriptors, 
where each element \(\mathbf{A}_{i,v}\) measures the similarity between the \(i\)-th vertex of the augmented mesh and the \(v\)-th vertex of the template mesh. 
Let \(\mathcal{M}_\text{sym}\) be a binary mask indicating vertex pairs that belong to symmetric body parts. 
We penalize similarity between such symmetric regions as:
\begin{equation}
\mathcal{L}_\text{sym}
= \frac{1}{M}\sum_{i,v} 
\mathcal{M}_\text{sym}(i,v)\,\mathbf{A}_{i,v},
\end{equation}
encouraging descriptors of mirrored parts to remain discriminative.  

The overall training objective combines these terms with the geodesic correspondence field-weighted contrastive supervision:
\begin{equation}
\mathcal{L}_\text{total}
=  \lambda_\text{soft}\mathcal{L}_\text{soft}
+ \lambda_\text{part}\mathcal{L}_\text{part}
+ \lambda_\text{sym}\mathcal{L}_\text{sym}.
\end{equation}

\begin{table*}[t]
\vspace{-7mm}
\centering
\small
\resizebox{0.95\linewidth}{!}{
\begin{tabular}{llccccccc}
\toprule
 & & \multicolumn{6}{c}{\textbf{Error} $\downarrow$} & \textbf{Efficiency} $\downarrow$ \\
\cmidrule(lr){3-8}
\textbf{Category} & \textbf{Methods} & \textbf{FAUST} & \textbf{SCAPE} & \textbf{SHREC19} & \textbf{DT4D-Intra} & \textbf{DT4D-Inter} & \textbf{Mean Error} & \textbf{Time (s)} \\ 
\midrule
Deformation-based & NJF~\cite{aigerman2022neural} & 5.9 & 11.7 & 9.6 & 43.4 & 32.8 & 20.68 & \underline{4.2} \\
2D Distillation & Diff3f~\cite{dutt2024diffusion} & 16.3 & 18.2 & 20.6 & 20.6 & 30.3 & 21.20 & 628.9 \\
Hybrid & DenoisingFM~\cite{zhuravlev2025denoising} & \textbf{1.9} & \textbf{2.4} & 4.2 & \underline{5.5} & 16.8 & 6.16 & 37.0 \\
& DiffuMatch~\cite{pierson2025diffumatch} & \textbf{1.9} & 4.4 & \textbf{3.9} & \textbf{1.8} & \underline{8.6} & \textbf{4.12} & 142.8 \\
\rowcolor{gray!12}
\textbf{Multimodal Intrinsic} & \textbf{SGSoft (Ours)} & \underline{2.5} & \underline{2.9} & \underline{4.0} & 8.1 & \textbf{8.3} & \underline{5.16} & \textbf{1.7} \\
\bottomrule
\end{tabular}
}
\vspace{-1mm}
\caption{\textbf{Comparison of mean geodesic error and inference time on standard benchmarks.}
SGSoft achieves competitive accuracy across most benchmarks while being the fastest among all baselines. 
Notably, it does not require pre-alignment, optimization, or post-refinement.}

\label{tab:benchmark_comparison}
\vspace{-3mm}
\end{table*}

\vspace{-3mm}
\subsection{Training and Inference}
\label{sec:method:train_infer}

During training, the pipeline is optimized using AdamW~\cite{loshchilov2017decoupled} under a combination of geodesic correspondence field-weighted, part, and symmetry losses within a curriculum learning scheme~\cite{bengio2009curriculum}.
Further implementation details regarding the training data curation, training architecture, and optimization settings are provided in the Supp.~B.

At inference, a pair of meshes are mapped to the multimodal intrinsic correspondence space to produce normalized per-vertex descriptors \(\hat{z}(x)\) through a single feed-forward pass.
Dense correspondences are subsequently retrieved via nearest-neighbor search in the descriptor space based on cosine similarity:
\begin{equation}
\vspace{-1mm}
\mathrm{corr}(x_i)
= \arg\max_{t_v}\, \hat{z}_\text{src}(x_i)^\top \hat{z}_\text{tgt}(t_v),
\vspace{-1mm}
\end{equation}
where \(\hat{z}_\text{src}(x_i)\) and \(\hat{z}_\text{tgt}(t_v)\) denote the normalized fused descriptors 
of the source and target vertices, respectively.  
This enables a fully feed-forward correspondence retrieval, without pre-alignment, pairwise optimization, and post-refinement. 

\section{Experiments}
\label{sec:experiments}

\subsection{Experimental Settings}
\paragraph{Benchmark Setup.}
We evaluate SGSoft on widely used correspondence benchmarks, including FAUST~\cite{bogo2014faust}, SCAPE~\cite{anguelov2005scape}, SHREC19~\cite{melzi2019shrec}, and DT4D (Intra/Inter)~\cite{li20214dcomplete}.  
For FAUST, SCAPE, and SHREC19, we use their remeshed variants to assess robustness under discretization and pose variations.  
DT4D datasets are employed to evaluate zero-shot inter-domain generalization across dynamic and non-human object categories.  

\paragraph{Accuracy and Efficiency Metrics.}
We report correspondence accuracy using the \emph{mean geodesic error} between predicted and ground-truth matches, averaged over all vertices and shape pairs.  
Efficiency is measured as wall-clock time for a representative shape pair, including both network input processing and correspondence retrieval, under a single-GPU configuration (RTX A6000). 

\paragraph{Baselines.}
Our evaluation compares correspondence methods across four paradigms: deformation, functional maps, large 2D vision models, and hybrids. We consolidate these baselines into three comparison categories: (1) deformation-based, (2) distillation from large 2D vision models, and (3) hybrid functional-map methods augmented with 2D semantic cues. More details for baselines are provided in Supp.~C.

\subsection{Baselines Comparison}
\paragraph{Near-Isometric Matching.}
We first evaluate SGSoft on near-isometric benchmarks, including FAUST, SCAPE and SHREC19, which assess correspondence stability under pose variation and remeshing.  
As shown in Table~\ref{tab:benchmark_comparison}, SGSoft attains accuracy comparable to the strongest refinement based and optimization based methods, despite not relying on any post-refinement or pairwise optimization. 
It reports mean geodesic errors of 2.5, 2.9 and 4.0 on FAUST, SCAPE and SHREC19, respectively, matching the performance of DenoisingFM~\cite{zhuravlev2025denoising} and DiffuMatch~\cite{pierson2025diffumatch} while operating in a purely feed forward manner.

\begin{figure}[t]
    \centering
    \includegraphics[width=0.6\linewidth]{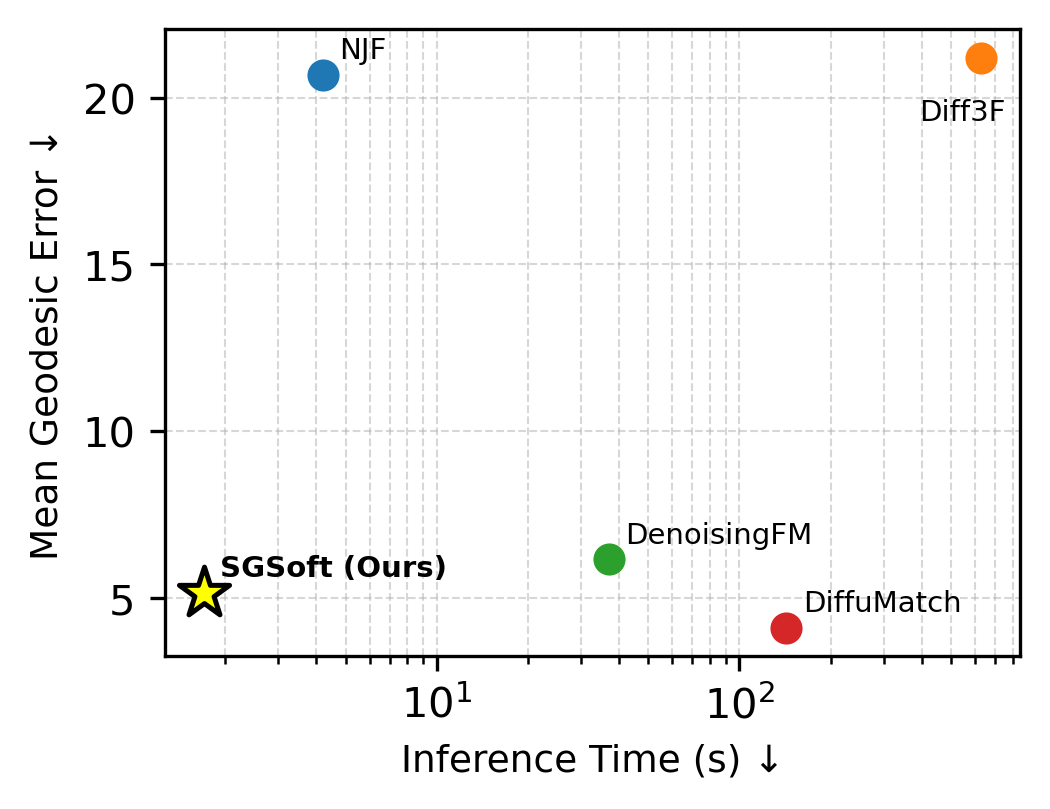} 
    \vspace{-5mm}
    \caption{
        \textbf{Trade-off between accuracy and efficiency.}
        SGSoft achieves the most balanced performance among all zero-shot methods.
    }
    \vspace{-5mm}
    \label{fig:table_balance}
\end{figure}

\begin{figure*}
    \noindent 
    \hspace{0.1\textwidth}
    \hspace{-1.7cm}
    \includegraphics[width=1.0\textwidth]{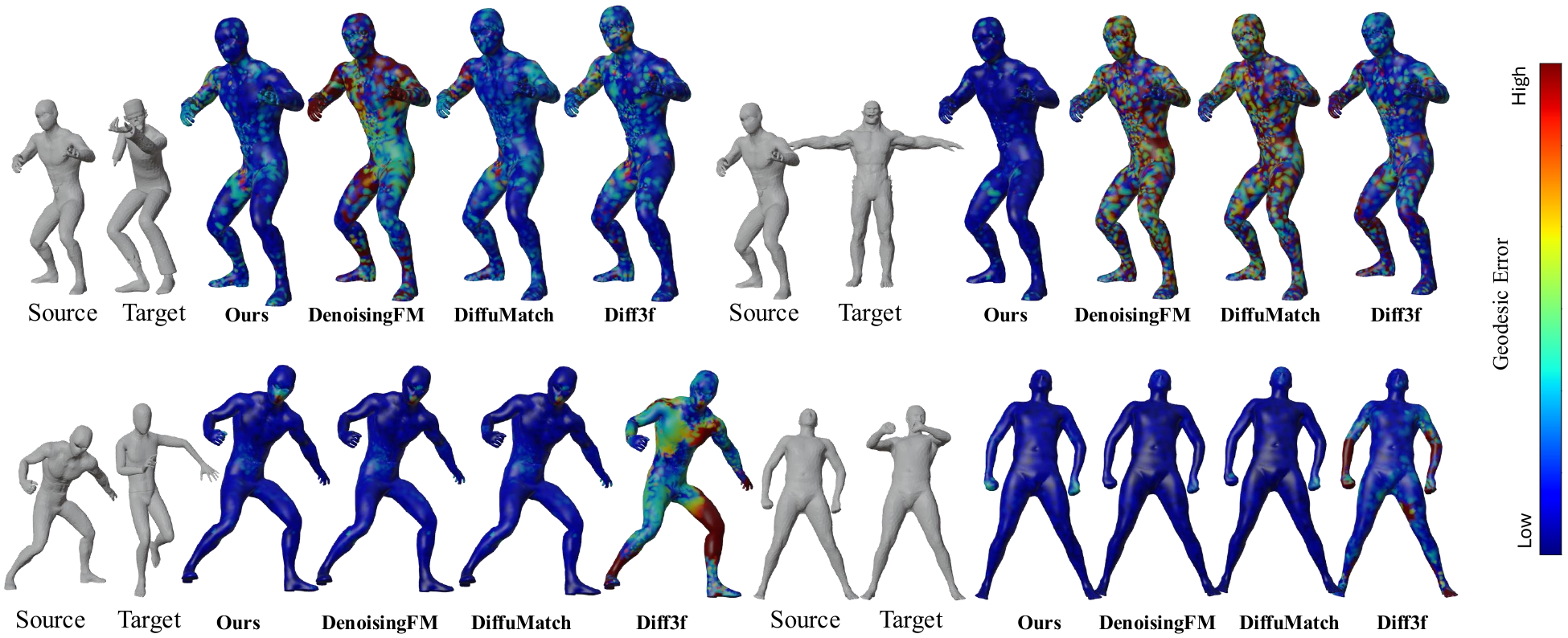}
    
    \vspace{-1mm}
    
    \par 
    \centering
    \caption{
\textbf{Qualitative comparison of correspondence accuracy.}
Per-vertex geodesic error is visualized from low (blue) to high (red).
SGSoft produces smooth and geodesically coherent error fields, 
whereas prior methods exhibit fragmented artifacts,  
especially under cross-category variation and symmetric ambiguity. Additional qualitative results are shown in Supp.~D. and ~E.
}

    \label{fig:qualitative}
\end{figure*}

\paragraph{Zero-Shot Cross-Domain Generalization.}
We next evaluate on DT4D-Intra and DT4D-Inter, which include dynamic and non-human object categories to assess generalization beyond human shapes Table~\ref{tab:benchmark_comparison}.
SGSoft delivers consistently strong performance across both settings, achieving mean geodesic errors of 8.1 and 8.3, respectively.
In contrast, DenoisingFM~\cite{zhuravlev2025denoising}, Diff3f~\cite{dutt2024diffusion} and NJF~\cite{aigerman2022neural} exhibit substantial degradation when transferred to non-human domains, indicating limited cross-category robustness.
While DiffuMatch~\cite{pierson2025diffumatch} attains high accuracy on DT4D-Intra through per-pair optimization, its performance drops notably on DT4D-Inter, suggesting weaker generalization across shape categories.
These results demonstrate that SGSoft preserves intrinsic geometric consistency and semantic alignment even under significant structural and domain variation.
\vspace{-3mm}
\paragraph{Accuracy–Efficiency Balance.}
While hybrid methods such as DenoisingFM and DiffuMatch achieve strong accuracy, they incur substantial inference time, ranging from 37 to 143 seconds per shape pair.  
As shown in Fig.~\ref{fig:table_balance}, SGSoft achieves the best trade-off between accuracy and efficiency, ranking within the top two across most benchmarks while operating at only 1.7 seconds per pair.  
This runtime is more than an order of magnitude faster than all baselines that rely on refinement or optimization, making SGSoft the only method capable of near real-time correspondence estimation with a single forward pass.
\vspace{-2mm}
\paragraph{Qualitative Comparison.}
Figure~\ref{fig:qualitative} visualizes correspondence quality by rendering the geodesic error between predicted matches and source ground-truth geodesic distances. 
Blue corresponds to the minimum error (0.0) and red to the maximum error within each visualization.  
The top row and bottom-left present DT4D-Inter results with the error range capped at 0.3, while the bottom-right presents SCAPE results with the range capped at 0.2.  
SGSoft produces spatially smooth and geodesically coherent error fields across diverse poses and categories, whereas the baselines exhibit fragmented artifacts, including instability under strong articulation (DenoisingFM, DiffuMatch) and confusion in symmetric regions (Diff3f).  
These qualitative observations are consistent with the quantitative findings and validate the robustness of our  multimodal descriptor.

\subsection{Ablation and Analysis}
\label{sec:experiments:analysis}

\paragraph{Ablation Studies and Component Analysis.}
We analyze the contribution of each component in SGSoft (Table~\ref{tab:ablation}) on benchmarks with diverse pose variations (SCAPE, SHREC19) and inter-category articulation (DT4D-Inter).
Removing the geodesic correspondence field (w/o $\tilde{S}$) and reverting to a standard InfoNCE loss formulation using hard anchors degrades performance, confirming the importance of the local continuous and topology-consistent supervision provided by our geodesic correspondence field.
Replacing the geodesic-aware contrastive loss with direct MSE regression (w/o $\tilde{S}$ Contrastive Loss) causes a drastic drop (e.g., 2.9→18.4 on SCAPE), showing that contrastive optimization is crucial for learning geodesically discriminative embedding learning.
Eliminating geodesic grouping and ungrouping (w/o Geodesic Grouping \&\ Ungrouping) reduces accuracy due to semantic leakage across adjacent regions, while removing the geodesic terms (w/o Geodesic Encoding) altogether weakens intrinsic consistency.
Lastly, omitting the symmetry loss (w/o Symmetry Loss) increases confusion between mirrored parts.
Together, these results demonstrate that geodesic modeling and the combination of geodesic correspondence field and contrastive learning are key to achieving stable and discriminative correspondence (see visualization of ablation in Supp.~F).
\begin{table}[t]
\centering
\caption{\textbf{Ablation study on key components of SGSoft.} 
Removing each component degrades performance across benchmarks, confirming their complementary contributions. 
All numbers denote mean geodesic error (\%) (lower is better).}
\vspace{-2.5mm}
\resizebox{0.9\linewidth}{!}{
\begin{tabular}{lccc}
\toprule
\textbf{Ablation Settings} & \textbf{SCAPE} & \textbf{SHREC19} & \textbf{DT4D-Inter} \\
\midrule
w/o $\tilde{S}$               & 3.7  & 6.6  & 10.7 \\
w/o $\tilde{S}$ Contrastive Loss  & 18.4 & 17.9 & 21.2 \\
w/o Geodesic Grouping \& Ungrouping  & 4.0  & 6.4  & 10.4 \\
w/o Geodesic Encoding                        & 3.3  & 4.9 & 9.3  \\
w/o Symmetry Loss                    & 7.8  & 8.2  & 15.6 \\
\midrule
\textbf{Full (SGSoft)}               & \textbf{2.9} & \textbf{4.0} & \textbf{8.3} \\
\bottomrule
\end{tabular}
}
\label{tab:ablation}
\vspace{0mm}
\end{table}


\begin{figure}[t]
    \vspace{-3mm}
    \centering    \includegraphics[width=\linewidth]{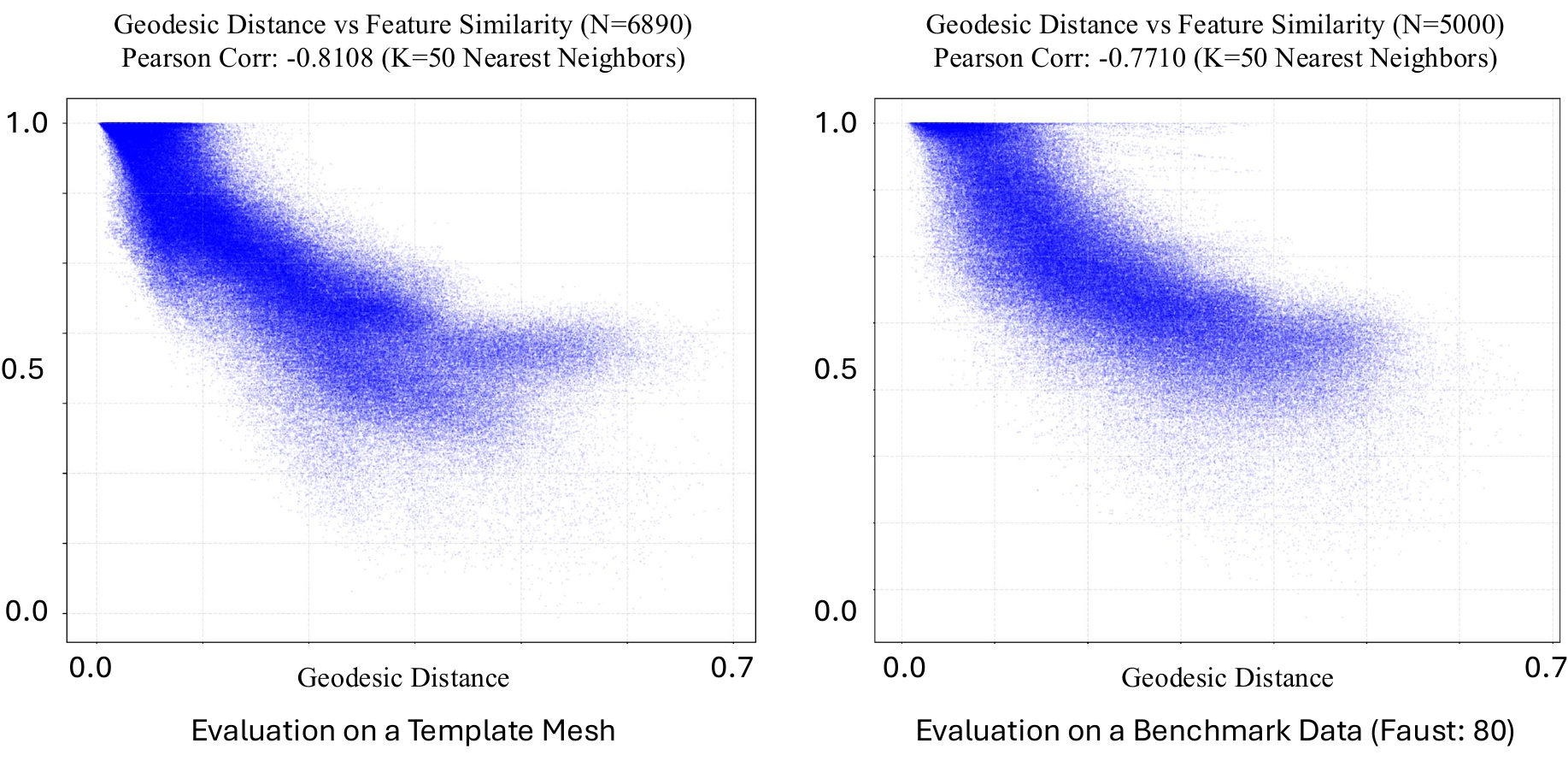}
    \caption{\textbf{Local Geodesic Consistency of the Descriptor Embedding.} 
    Analysis of the K=50 nearest geodesic neighbors shows a strong negative correlation between feature similarity and distance on the template (left, Pearson: -0.81) and a benchmark mesh (right, Pearson: -0.77), confirming high fidelity to local geometry.
    }
    \vspace{-6mm}
    \label{fig:geo_corr}
\end{figure}

\vspace{-3mm}
\paragraph{Geodesic Consistency of the Learned Embedding.}
We assess the surface-aware locality encoded by SGSoft by examining the correlation between descriptor similarity and intrinsic geodesic distance.
Our training objective $\mathcal{L}_\text{soft}$ uses a localized, geodesic correspondence field that enforces consistency only within adaptively sampled intrinsic neighborhoods (see Supp.~A.1).  
To evaluate how well the learned descriptor's embedding reflects this supervisory structure, we compute the correlation within the $K=50$ nearest geodesic neighbors for each vertex.
As shown in Fig.~\ref{fig:geo_corr}, which plots feature similarity (Y-axis) against geodesic distance (X-axis), the embedding exhibits a strong and smooth negative correlation (Pearson: -0.81 on the template mesh and -0.77 on a benchmark mesh).  
This indicates that SGSoft preserves intrinsic geometry with high fidelity at local scales (as seen in local detail preservation in Fig.~\ref{fig:teaser}).  
The monotonic decay closely mirrors the structure imposed by the geodesic correspondence field supervision, confirming that the model learns a robust and geodesically consistent feature representation within these neighborhoods.

\vspace{-3mm}
\paragraph{Multimodal Representations and Downstream Consistency.}
The soft multimodal feature representation learned by SGSoft is not only robust to geometric and semantic variations but also to changes in the underlying 3D representation itself, as illustrated in Fig.~\ref{fig:teaser}. This representation also enables reliable transfer to downstream tasks such as semantic segmentation (Fig.~\ref{fig:teaser}) and deformation transfer. See more results in Supp.~I.

\section{Conclusion}
\label{sec:conclusion}

We introduced SGSoft, a unified framework for generalizable and efficient dense 3D correspondence.  
By integrating a geodesic correspondence field with a multimodal fusion descriptor, SGSoft achieves geometric-fidelity, semantic precision, and real-time inference without pre-alignment or post-processing.  
Leveraging semantic priors from Uni3D enables robust cross-domain generalization, while the geodesic formulation ensures surface-consistent and deformation-robust mapping. 

This study highlights the potential of combining intrinsic geometry with pre-trained semantic knowledge to build a scalable, interpretable, and foundation-ready paradigm for 3D correspondence learning. This finding also illuminates a critical trade-off between semantic generalization and geometric specificity in current 3D foundation models, highlighting a key challenge for the next generation of 3D representations.

\clearpage
\setcounter{page}{1}
\maketitlesupplementary

\appendix
\section{Extended Method Details}
\label{sec:minor}

\input{}
\subsection{Canonical Geodesic Correspondence Field Construction}
The geodesic correspondence field $\tilde{S}$ serves as the supervisory signal of our framework.
It provides stable, continuous, and topology-invariant correspondence likelihoods defined entirely on the canonical template mesh.  
This field is computed \emph{once} on the template and reused across all augmented meshes, enabling consistent supervision under arbitrary remeshing, while avoiding repeated geodesic computation on each variant.

\begin{figure*}[t]
    \centering
    \includegraphics[width=\linewidth]{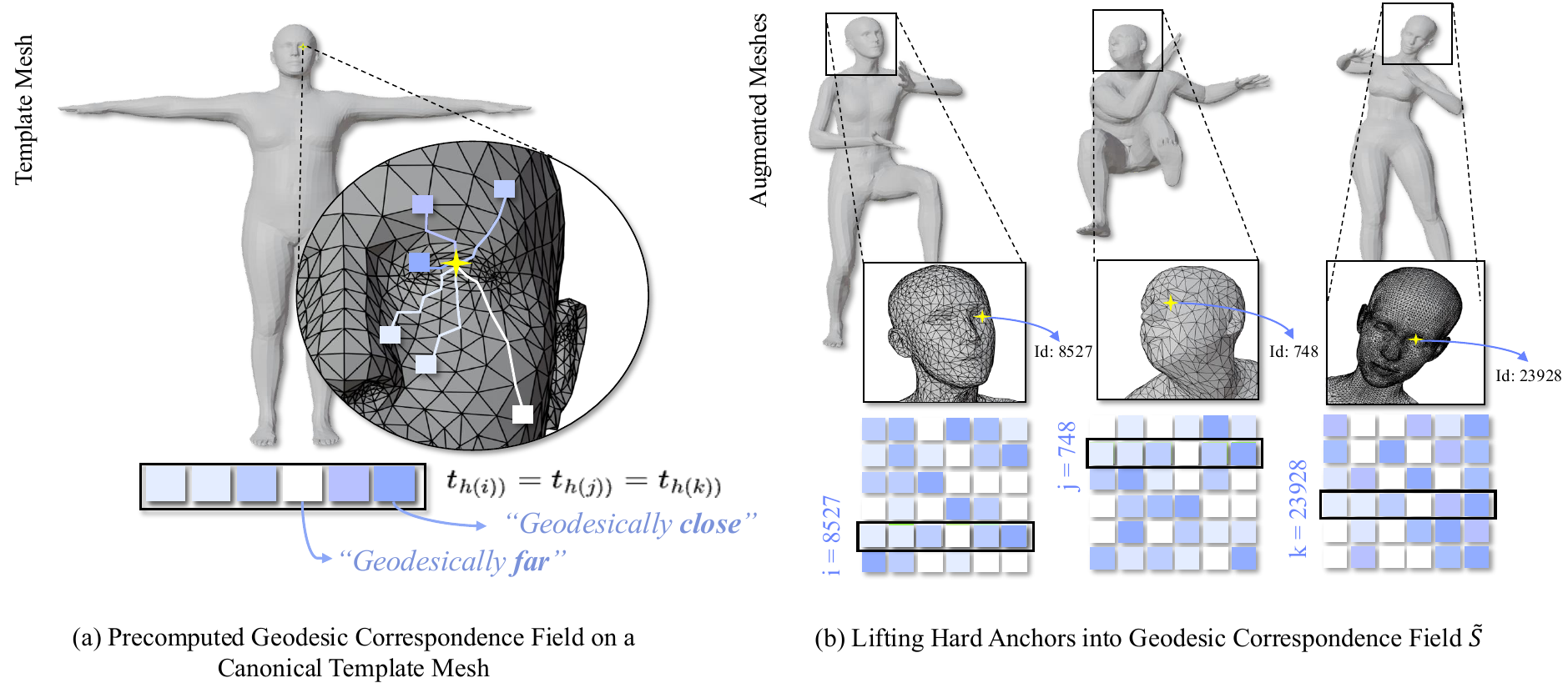}
    \caption{\textbf{Topology- and deformation-consistent geodesic correspondence field.}
    (a) We precompute a continuous geodesic correspondence field on a canonical template mesh,
    where each hard anchor is lifted to an intrinsic geodesic neighborhood defined by surface distances.
    (b) The field is consistently reused across remeshed and deformed shapes via hard anchor mapping,
    yielding a topology- and deformation-invariant correspondence field $\tilde{S}$.}
    \label{fig:supp_geodesic_field}
\end{figure*}

We denote the canonical template mesh as $T = (V_T, F_T)$ with vertex set $V_T = \{t_1, \dots, t_{N_T}\}$.  
For each augmented mesh vertex $x_i$, a one-to-one mapping $h(i)$ is maintained through the augmentation chain, linking $x_i$ back to its canonical vertex $t_{h(i)}$.  
This mapping allows the precomputed template field $\tilde{S}_{h(i), v}$ to be directly used as the correspondence distribution for $x_i$, eliminating the need for geodesic recalculation on augmented meshes. (See Fig.~\ref{fig:supp_geodesic_field})

\paragraph{Step 1. Intrinsic Precomputation on the Template.}
We first compute a set of intrinsic geometric quantities on the canonical template mesh:
\begin{itemize}
    \item local vertex area $A[v]$ is computed as one-third of the total area of its incident one-ring faces.
    \item global median vertex area $a_{\mathrm{med}}$.
    \item curvature estimate $\kappa[v]$ (e.g., mean curvature magnitude ~\cite{rusinkiewicz2004estimating}).
    \item geodesic distances $d_{\mathrm{geo}}(p, v)$ for all pairs of vertices on $T$, computed using an intrinsic solver (e.g., heat method~\cite{crane2013geodesics}).
\end{itemize}

\paragraph{Step 2. Adaptive Neighborhood Size.}
For each template vertex $t_i$, we compute a density coefficient inversely proportional 
to its local surface area:
\begin{equation}
    \rho_i = \frac{a_{\mathrm{med}}}{A[i] + \epsilon},
\end{equation}
where $A[i]$ is the vertex area and $a_{\mathrm{med}}$ is the global median area.
The density $\rho_i$ is clipped to a predefined range and translated into an adaptive
neighborhood size:
\begin{equation}
    K_i = \mathrm{clip}\!\left(
    \mathrm{round}\!\left(K_{\mathrm{base}} \cdot \rho_i^{\alpha}\right),
    K_{\min}, K_{\max}
    \right).
\end{equation}
We set $K_{\mathrm{base}}=32$, $K_{\min}=8$, and $K_{\max}=50$.

This strategy allocates denser neighborhoods in geometrically detailed regions
(e.g., fingers, facial features) and avoids redundant sampling over flat areas,
yielding a balanced intrinsic support.
By additionally capping the neighborhood size, we maintain computational efficiency and sparsity,
while preventing over-dense connections in locally uniform regions.

\paragraph{Step 3. Base Geodesic Kernel.}
Given the precomputed geodesic distance matrix, we define an intrinsic neighborhood
\begin{equation}
    \mathcal{N}_i^{(K_i)} =
    \text{Top-}K_i \;\text{vertices } t_v
    \;\text{with smallest } d_{\mathrm{geo}}(t_i, t_v).
\end{equation}
For each $v \in \mathcal{N}_i^{(K_i)}$, we construct a geodesic Gaussian kernel:
\begin{equation}
    \hat{S}_{i,v}
    = \exp\!\left( -\frac{d_{\mathrm{geo}}(t_i,t_v)^2}{\sigma^2} \right),
\end{equation}
which represents the base intrinsic correspondence kernel
defined purely by geodesic distances on the canonical template surface.
This kernel is later modulated by geometric and semantic constraints to obtain
the final correspondence field $\tilde{S}$.

\paragraph{Step 4. Geometric and Semantic Modulation.}
We further refine the intrinsic field by incorporating geometric saliency
and semantic constraints.

\emph{Geometric saliency weighting.}
To emphasize high-curvature structures, we define a curvature-aware weight:
\begin{equation}
    A_{i,v} = 
    \left(1 + |\kappa_v|\right)
    \,\mathbb{1}\!\left[t_v \in \mathcal{N}_i^{(K_i)}\right],
\end{equation}
where $\kappa_v$ denotes the mean curvature magnitude at $t_v$. 

\emph{Semantic plausibility filtering.}
To suppress implausible correspondences (e.g., hand-to-foot),
we apply a same-part mask:
\begin{equation}
    \Pi_{i,v} = \mathbb{1}\!\left[\ell(t_v) = \ell(t_i)\right],
\end{equation}
where $\ell(\cdot)$ denotes the semantic part label of a vertex.

\paragraph{Step 5. Final Geodesic Correspondence Field.}
The geodesic-aware soft correspondence field is defined as:
\begin{equation}
    \tilde{S}_{i,v} \propto
    A_{i,v}
    \cdot
    \hat{S}_{i,v}
    \cdot
    \Pi_{i,v},
    \qquad
    \sum_v \tilde{S}_{i,v} = 1,
\end{equation}
This yields a compact, sparse, and intrinsically geometry-aware correspondence
distribution centered at vertex $t_i$.

\paragraph{Canonical Reuse Across Augmented Meshes.}
Since the correspondence field $\tilde{S}$ is defined entirely on the canonical
template using intrinsic geodesic distances, it is computed \emph{once} 
and reused for all augmented meshes.  
For an augmented vertex $x_i$ with canonical anchor $h(i)$, we directly assign
\begin{equation}
    \tilde{S}_{x_i}(v) = \tilde{S}_{h(i), v}.
\end{equation}
This provides consistent, topology-invariant supervision under remeshing and deformation, while completely eliminating the need for
recomputing geodesic distances on each augmented shape.

\subsection{Hierarchical Semantic Aggregation}
To integrate semantic information across multiple transformer depths, we adopt a hierarchical aggregation strategy that fuses intermediate features from selected layers of the Uni3D~\cite{zhou2023uni3d} backbone. 
Shallow layers capture fine geometric details, while deeper layers encode global semantic context. 

Specifically, for the 40 layers of Uni3D transformer backbone, we extract features from layers 
$\ell = \{8,16,24,39\}$ and perform hierarchical fusion over these representations.

Given features $\{H^{(\ell)}\}_{\ell \in \mathcal{I}}$ from selected layers $\mathcal{I}$, we project them into a shared space and aggregate them with learned layer weights:
\begin{equation}
f_\mathrm{sem} =
\sum_{\ell \in \mathcal{I}} \alpha_\ell\, H^{(\ell)},
\quad
\alpha_\ell =
\mathrm{softmax}(a_\ell).
\end{equation}
This weighted fusion balances contributions from low- and high-level features, producing a unified semantic embedding that captures discriminative semantics and global contextual structure.

\subsection{Ablation Studies}
We analyze the contribution of three core components, namely
(A) adaptive neighborhood size (density-aware sampling),
(B) geometric saliency weighting, and
(C) hierarchical semantic aggregation.
As shown in Table~\ref{tab:supp_ablation}, removing any of these components degrades performance across benchmarks.

w/o A significantly increases the error on SCAPE and SHREC19, highlighting the importance of density-aware intrinsic sampling.
In contrast, DT4D-Inter contains large global shape variations across subjects. In such settings, a uniform neighborhood leads to more stable global matching, whereas density-aware sampling, which is optimized for local isometric detail, can introduce an unnecessary locality bias.

w/o B leads to the largest degradation on SCAPE and SHREC19, demonstrating that curvature-based saliency is essential for dynamically posed and highly articulated structures.

w/o C also causes substantial degradation on all benchmarks, indicating that multi-level semantic fusion is crucial for robust part discrimination and consistent semantic alignment across large pose and shape variations.

Overall, the full model (A+B+C) achieves the best performance, and each component provides
a complementary benefit to dense correspondence accuracy.
\begin{table}[h]
\centering
\caption{\textbf{Ablation study on three extended core components of SGSoft.}
Reported values are mean geodesic error ($\downarrow$) across SCAPE, SHREC19, and DT4D-Inter.}
\label{tab:supp_ablation}

\resizebox{\linewidth}{!}{
\begin{tabular}{l|cccc}
\toprule
Variant & SCAPE & SHREC19 & DT4D-Inter & Avg. \\
\midrule
\textbf{SGSoft (full; A+B+C)}
& \textbf{2.9} & \textbf{4.0} & 8.3 & \textbf{5.03} \\
\midrule
w/o A (Uniform neighborhood)
& 7.1 & 8.7 & \textbf{8.0} & 7.93 \\
w/o B (No geometric saliency)
& 8.0 & 11.9 & 8.3 & 9.40 \\
w/o C (No hierarchical aggregation)
& 4.8 & 12.5 & 11.5 & 9.60 \\
\bottomrule
\end{tabular}
}
\end{table}

\section{Training Dataset and Implementation Details}
\label{sec:impl}

\subsection{Training Dataset} 
\begin{figure}[t]
  \centering
  \includegraphics[width=\linewidth]{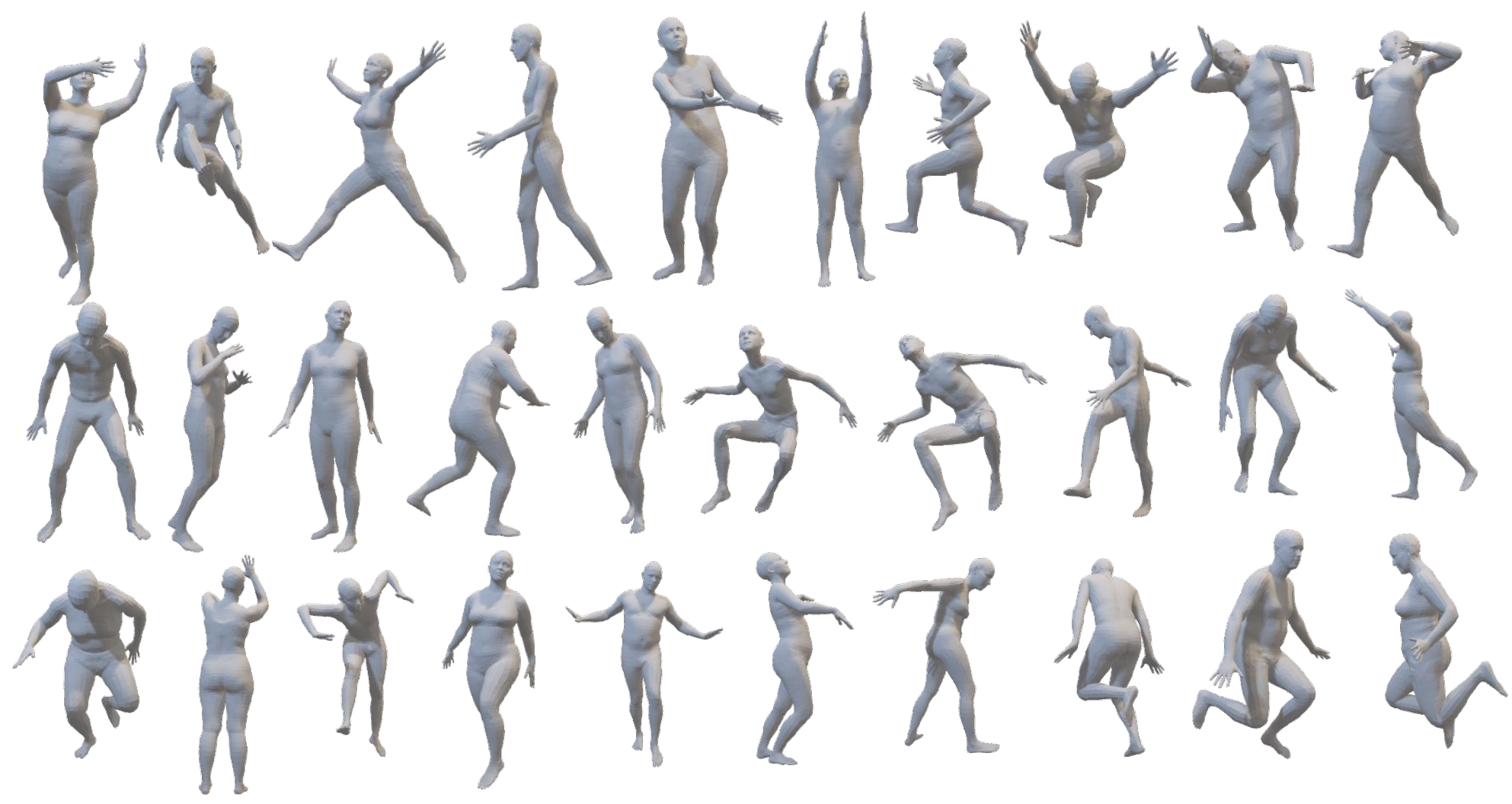}
  \caption{
    \textbf{Training dataset overview.}
    We sample 183 base human meshes from the SMPL model using VPoser-driven body poses and random body shapes, and generate 10 topologically perturbed variants for each base mesh, yielding 1{,}830 training meshes in total. 
  }
  \label{fig:supp_dataset}
\end{figure}

We construct our training set by randomly sampling body shapes and poses using the SMPL model~\cite{Loper2015}.
Pose parameters are generated in the VPoser latent space to ensure physically plausible and realistic human motions.
From this process, we obtain 183 base human meshes, as illustrated in Fig.~\ref{fig:supp_dataset}.
Each base mesh is further diversified with 10 topological variations, resulting in a total of 1,830 training meshes, each associated with a precomputed geodesic correspondence field.

\subsection{Implementation Details}

\emph{Uni3D Backbone.}
We use Uni3D with the EVA-Giant-Patch14-560 backbone (1B parameters) as our 3D semantic encoder.  
Each input point cloud is partitioned into $G=512$ local patches, each containing $M=64$ points, using Farthest Point Sampling with geodesic-aware assignment.

\emph{Curriculum Learning.}
We adopt a simplified two-phase curriculum learning strategy.
During a warm-up stage (1 epoch), training focuses solely on the part classification loss to stabilize semantic alignment.
This is followed by a transition stage of 8 epochs, during which classification and contrastive losses are smoothly blended using a cosine schedule.
The final loss weights are set to $(w_{\text{part}}, w_{\text{cont}}) = (0.6,\,0.3)$.
We additionally apply a symmetry consistency regularization with weight $0.3$.

\emph{Training.}
We train the model for a total of 13 epochs with batch size $1$.
Training takes approximately 17 hours on a single NVIDIA A6000 GPU (48GB).
To ensure reproducibility, the full codebase and trained models will be released.

\section{Baselines}
The selected baselines in main paper are drawn from related studies \textbf{where official codes are publicly available.}

\begin{enumerate}[label=(\roman*)]

\item \textbf{Deformation-based feed-forward.}
NJF~\cite{aigerman2022neural} learns a neural deformation field, predicting dense correspondences via forward regression within a canonical domain.

\item \textbf{2D-pretrained feature distillation.}
Diff3F~\cite{dutt2024diffusion} transfers 2D diffusion features to 3D surfaces via multi-view rendering conditioned on depth and normal maps.

\item \textbf{Hybrid with Functional Maps.}
DiffuMatch~\cite{pierson2025diffumatch} combines 2D diffusion features with iterative functional map refinement for dense alignment.
DenoisingFM~\cite{zhuravlev2025denoising} estimates correspondences via a denoising diffusion process that reconstructs functional maps from noisy spectral embeddings.

\end{enumerate}

For all baselines, we report the performance from their official implementations under default configurations, with two exceptions: for DenoisingFM~\cite{zhuravlev2025denoising}, we adopt the \(32\times32\) functional map resolution (its best on the DT4D benchmark) for consistency; and for NJF~\cite{aigerman2022neural}, we report the performance following the configuration from the recent baseline~\cite{pierson2025diffumatch}, as the original work does not provide an official configuration for this shape correspondence benchmark.

\section{Additional Qualitative Comparison}
\begin{figure}[t]
  \centering
  \includegraphics[width=\linewidth]{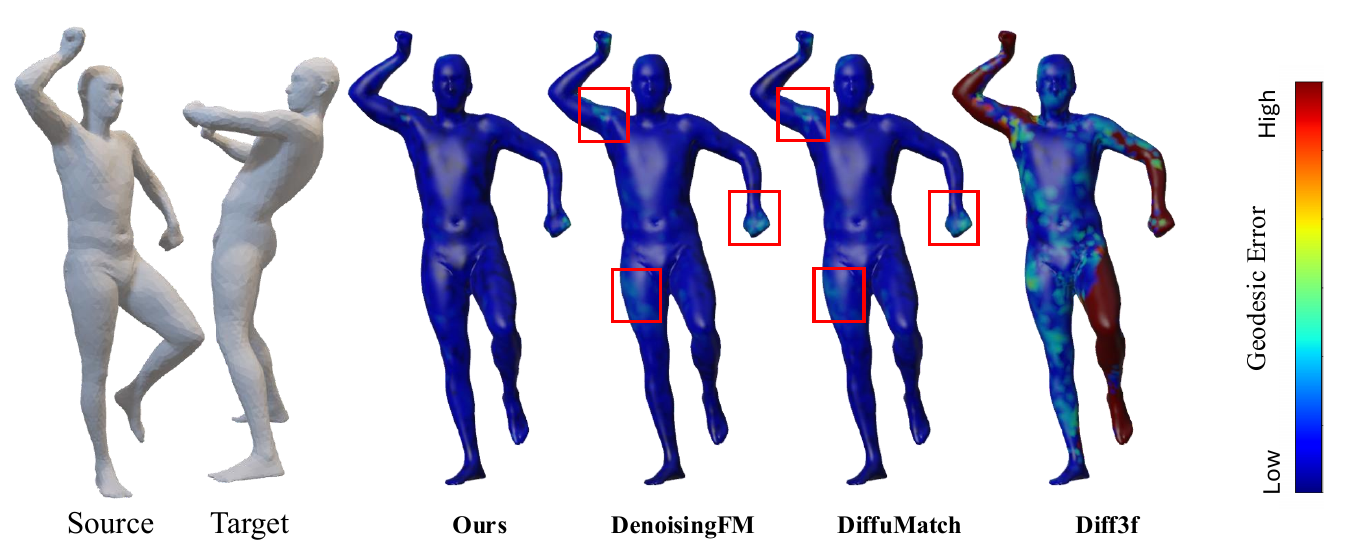}
  \caption{
  \textbf{Additional qualitative comparison.}
  Geodesic error visualization for challenging correspondence under dynamic pose deformation.
}

  \label{fig:supp_comparison}
\end{figure}

Figure~\ref{fig:supp_comparison} shows an additional qualitative comparison under a challenging scenario with dynamic pose deformations between the source and target shapes.
While functional map-based methods such as DenoisingFM and DiffuMatch handle large articulations reasonably well, they suffer from noticeable errors in locally ambiguous regions with minor deformations, such as the upper arm, hand, and thigh (highlighted in red boxes).

In contrast, SGSoft produces more stable and semantically consistent correspondences in these regions by leveraging the proposed geodesic correspondence field together with semantic cues.

\section{Additional Results}

\subsection{Symmetry Disambiguation}
Figure~\ref{fig:supp_symmetry_disambiguation} visualizes SGSoft’s ability to disambiguate 
left-right and front-back symmetries across FAUST, SCAPE, and SHREC19. 
For each example, we transfer correspondences from a source mesh to a target mesh under a different pose.
The uncolored source and target meshes are shown in their original poses,
while the colorized meshes are rotated to a canonical front view for clearer visualization.
SGSoft consistently assigns stable and semantically correct colors to corresponding parts
(e.g., left/right arms and legs) even under strong articulation.

\subsection{Zero-Shot Cross Domain Generalization}
Figure~\ref{fig:supp_zeroshot} presents zero-shot generalization results from human-like training shapes to stylized character meshes.  
The \emph{Intra} columns evaluate correspondence quality within the same character family, 
while the \emph{Inter} columns demonstrate transfer across different families with large variations in appearance and body proportions.  
Although SGSoft is trained exclusively on SMPL-like human bodies, 
it preserves dense correspondences even across highly non-isometric shapes.  
These results indicate that our multimodal, intrinsic representation generalizes beyond the training domain 
and remains robust under strong geometric and stylistic distribution shifts.
For clarity of comparison, the source and target meshes are shown in their original poses, 
while the colored correspondence results are rendered after a consistent front-facing rotation.

\subsection{Representation Robustness}

\begin{figure}[t]
  \centering
  \includegraphics[width=\linewidth]{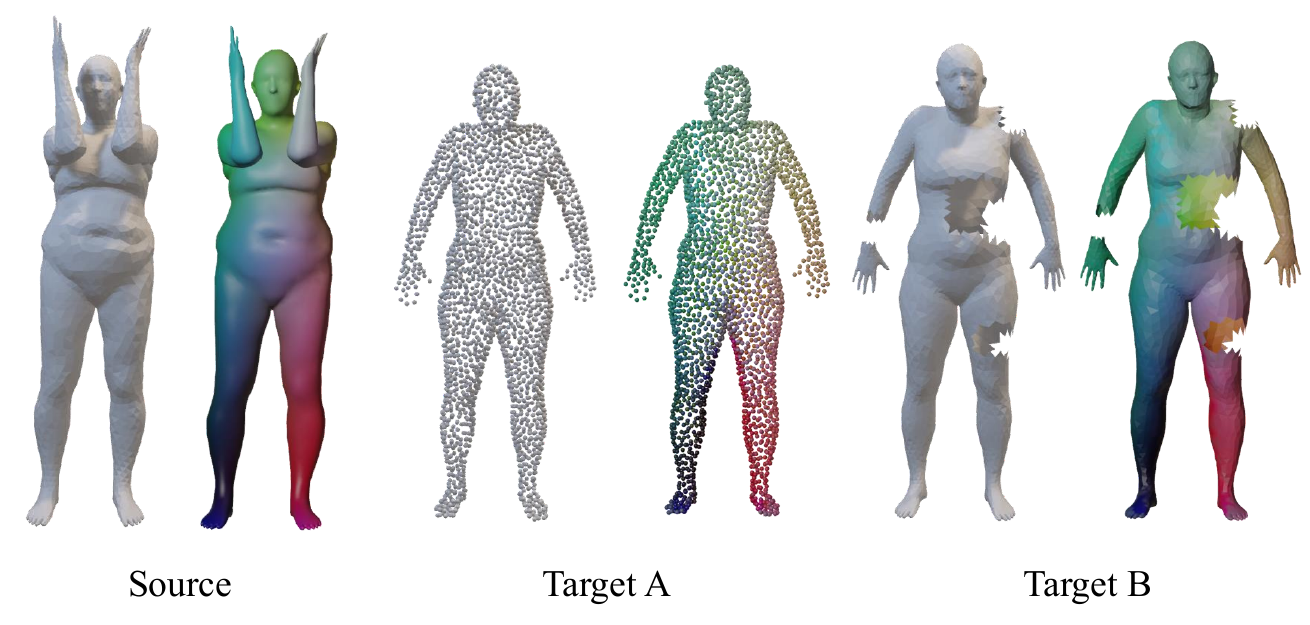}
  \caption{
    \textbf{Representation robustness.}
    Target~A is a point cloud and Target~B is a partial mesh.
    SGSoft produces consistent correspondence colors across both targets,
    showing representation-agnostic behavior.
  }
  \label{fig:supp_representation}
\end{figure}

In Figure~\ref{fig:supp_representation}, we probe the robustness of the learned multimodal intrinsic space across different geometric representations and levels of completeness. We match a single source to two structurally different targets: \emph{Target~A}, represented as a point cloud, and \emph{Target~B}, given as a partial mesh. Despite these variations, SGSoft assigns consistent semantic correspondences across both targets, indicating stability under moderate structural changes.

\begin{figure}[t]
  \centering
  \includegraphics[width=\linewidth]{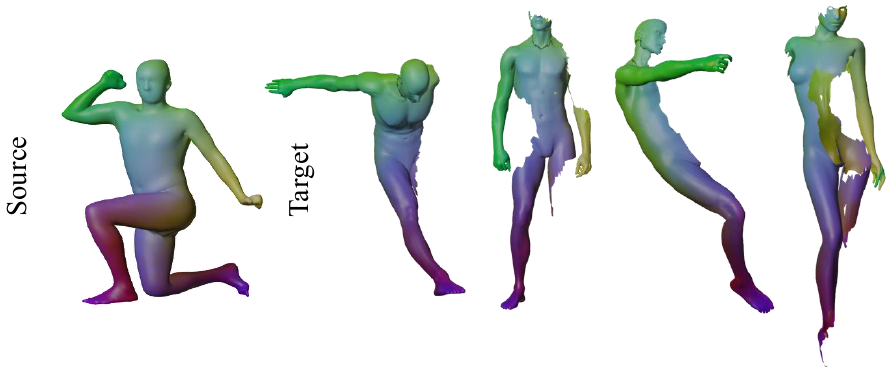}
  \caption{
    \textbf{Qualitative results on samples from SHREC'16.}
    Despite disconnected geometry, SGSoft maintains semantically consistent correspondences in visible areas. 
  }
  \label{fig:supp_partial}
\end{figure}

To further evaluate robustness under more severe incompleteness, we additionally present qualitative results on the SHREC'16 partial shape benchmark (Fig.~\ref{fig:supp_partial}). Even with large missing regions and disconnected geometry, SGSoft maintains semantically coherent correspondences in most visible areas, although performance degrades in extreme cases in Fig.~\ref{fig:supp_failure}. These results suggest that the learned descriptor remains robust across a spectrum of geometric representations and partial observations.

\subsection{Topology Robustness}
To evaluate robustness under remeshing, Fig.~\ref{fig:supp_topology} visualizes correspondence results with varying vertex resolutions and triangulations.
Despite large changes in connectivity and sampling density, SGSoft preserves smooth and semantically consistent color patterns across a wide range of topologies.
These results suggest that our template-guided geodesic correspondence field provides largely topology-invariant supervision that generalizes across different surface discretizations in practice. 
Table~\ref{tab:topology_random} further quantifies this behavior on 10 randomly sampled SHREC19 pairs with 50 random remeshing perturbations in total.
Although the average geodesic error slightly increases from 0.0325 to 0.0386 after remeshing, the overall performance remains in a comparable range, supporting the practical robustness of SGSoft under substantial changes in mesh resolution and triangulation.

\begin{table}[h]
\centering
\begin{minipage}{0.38\textwidth}
\centering
\footnotesize
\caption{Geodesic correspondence error under random remeshing on SHREC19. Lower is better.}
\label{tab:topology_random}
\begin{tabular}{lc}
\toprule
Setting & Mean Error $\downarrow$ \\
\midrule
Baseline & 0.0325 \\
Avg. remeshing & 0.0386 \\
\bottomrule
\end{tabular}
\end{minipage}
\end{table}

\subsection{Analysis of Multimodal Descriptor}

\begin{figure}[t] \centering \includegraphics[width=\linewidth]{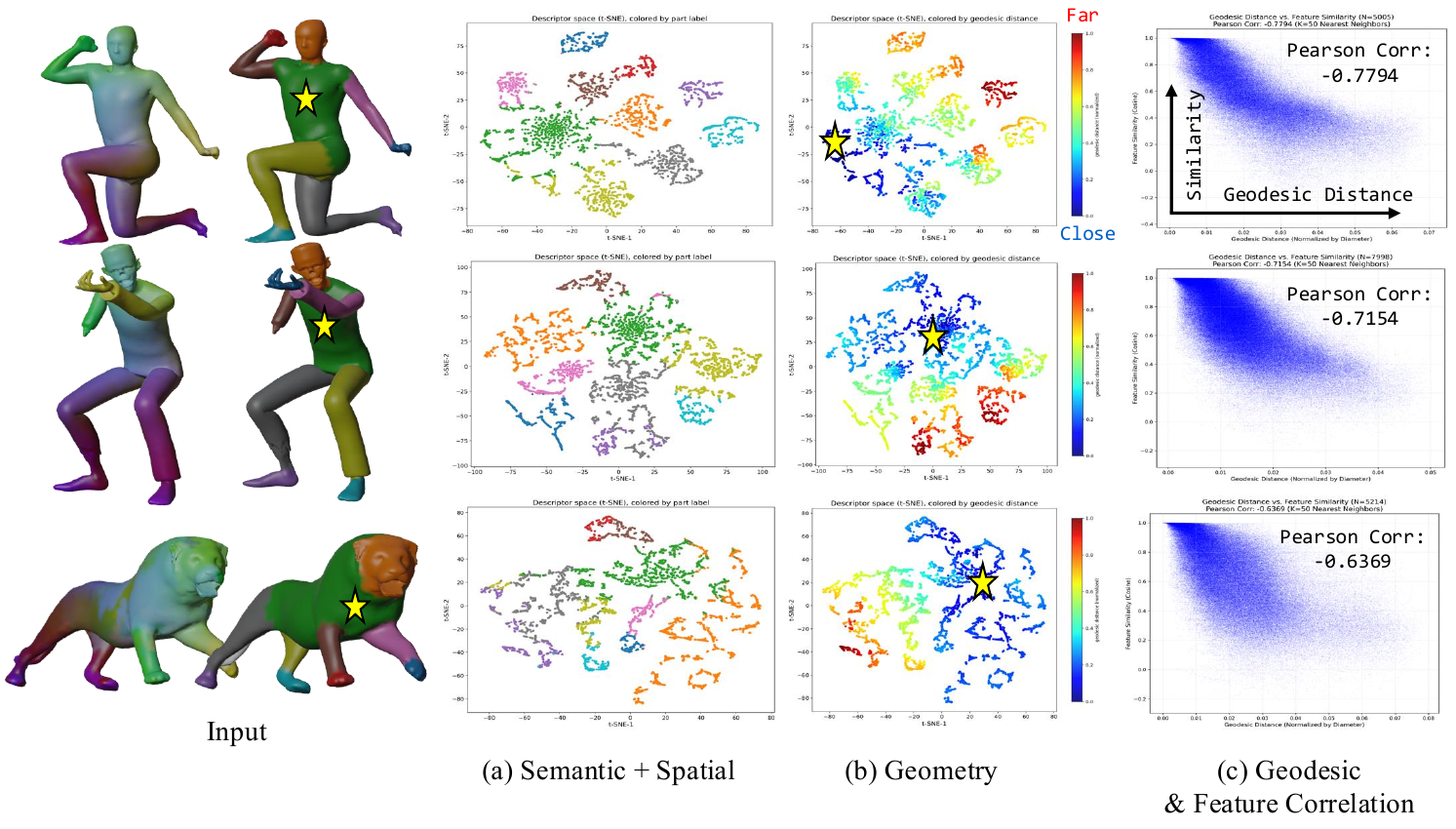} \caption{\textbf{t-SNE visualization and feature–geometry correlation.}
(a) The descriptor space forms semantic clusters and separates symmetric regions across diverse shapes.
(b),(c) Feature similarity from a reference point (star) negatively correlates with geodesic distance, consistent with Pearson trends.}\label{fig:supp_tsne} \end{figure}
To demonstrate the distinct roles of each modality, we analyze the structure of the learned feature space using t-SNE and feature–geometry correlation. As shown in Fig.~\ref{fig:supp_tsne}(a), the descriptor space exhibits clustering patterns aligned with semantic body parts, while also separating symmetric regions, indicating effective semantic and spatial discrimination. These patterns remain consistent across diverse shape categories, including humans, stylized characters, and animals, suggesting that the learned representation generalizes beyond the training domain.

Furthermore, feature similarity from a reference point (denoted by a star) shows a clear negative correlation with geodesic distance (Fig.~\ref{fig:supp_tsne}(b),(c)), which is also reflected in the Pearson correlation trends. This indicates that the representation captures intrinsic geometric structure.

Overall, these observations suggest that SGSoft integrates semantic, geometric, and spatial cues into a unified feature space, where semantic priors from Uni3D contribute to global alignment, and the geodesic correspondence field encourages local consistency.

\section{Visualization of Ablation Studies}

\begin{figure*}
    \vspace{-6mm}
    \makebox[\textwidth][c]{%
        \includegraphics[width=0.9\textwidth]{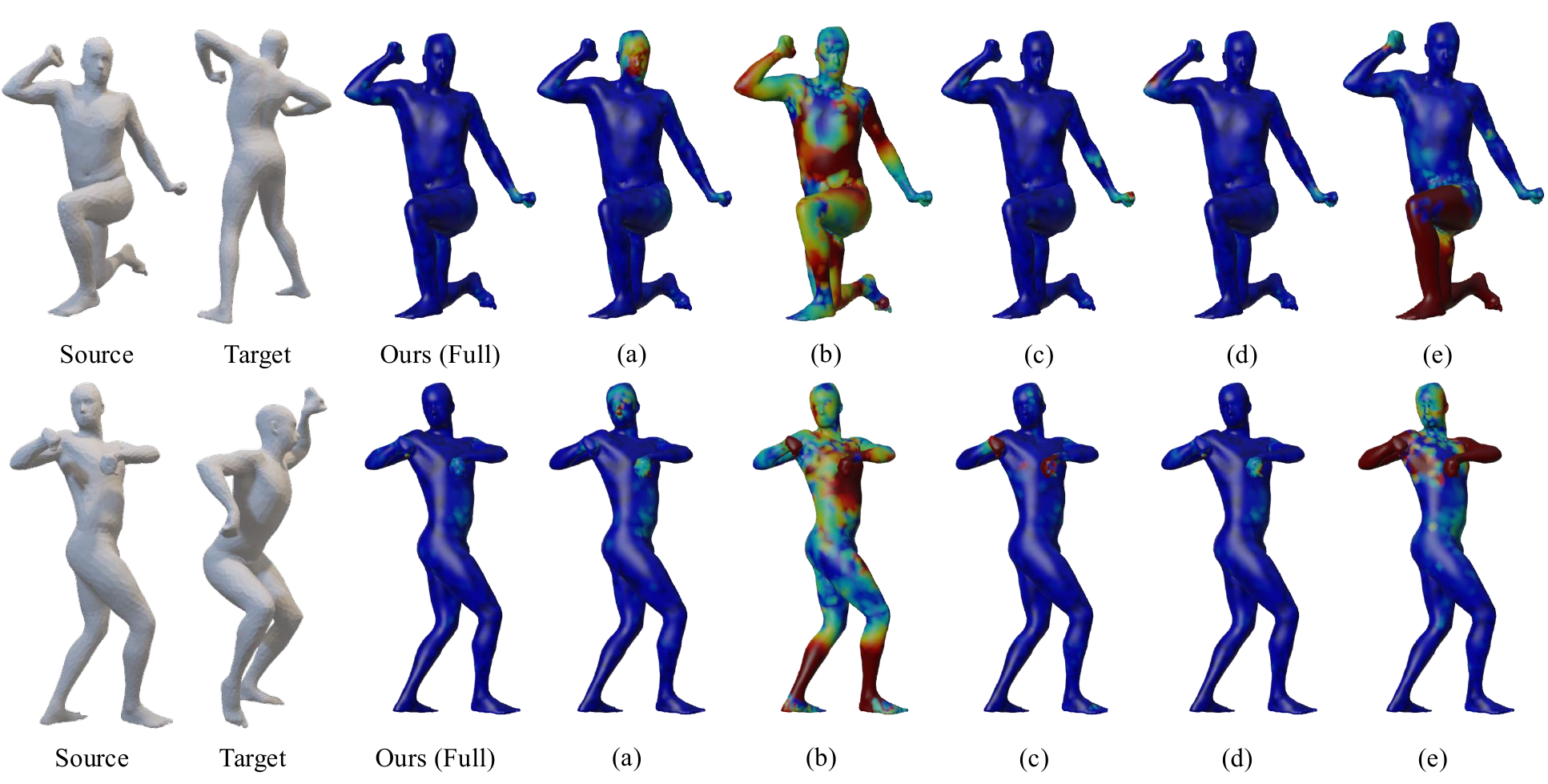}
    }
    \vspace{-5mm}
    \captionof{figure}{
      \textbf{Visualization of ablation studies.}
      From left to right: source mesh, target mesh, SGSoft (full), and ablated models
      (a) w/o geodesic correspondence field $\tilde{S}$,
      (b) w/o contrastive loss on $\tilde{S}$,
      (c) w/o geodesic grouping \& ungrouping,
      (d) w/o geodesic encoding, and
      (e) w/o symmetry loss.
      Removing each component produces visible artifacts such as color bleeding across parts, symmetry flips, and misalignment near joints, whereas the full model yields sharp and part-consistent correspondences.
    }
    \vspace{-5mm}
    \label{fig:supp_vis_ablation}
\end{figure*}

Figure~\ref{fig:supp_vis_ablation} presents qualitative results for the ablation experiments
reported in the main paper.
Removing the geodesic correspondence field $\tilde{S}$ (a) leads to noticeable errors in
locally structured areas such as the face, where fine geometric details are critical.
Disabling contrastive supervision (b) results in severe performance degradation and
inconsistent color transfer across parts.

Further removing geodesic grouping/ungrouping or geodesic encoding (c) causes significant
misalignment, often producing color bleeding between adjacent semantic regions.
Finally, removing the symmetry loss leads to frequent left-right flips in symmetric body
parts.

In contrast, the full SGSoft model produces sharp, part-consistent correspondences with
clear semantic boundaries, visually corroborating the quantitative trends reported in the ablation study of the main paper. 

\section{Preliminary Study on 3D Backbone Selection}
\subsection{Evolution of 3D Shape Representation}
\label{sec:backbone_evolution}

Recent advances in 3D shape representation learning have followed different directions depending on the input modality and output granularity.  
Early mesh-based methods directly operate on surface connectivity and geometry (e.g., MeshNet~\cite{feng2019meshnet}, MeshCNN~\cite{hanocka2019meshcnn}, DiffusionNet~\cite{sharp2022diffusionnet}). 
While these approaches preserve structural information, they often require heavy preprocessing and are less scalable for large-scale semantic learning.

Mesh representations can be readily converted into point clouds, and many recent works therefore focus on point-based architectures that offer higher flexibility and better scalability. 
Foundational models such as PointNet~\cite{qi2017pointnet} and PointNet++~\cite{qi2017pointnet++} introduced permutation-invariant and locality-aware feature extraction.  
Transformer-based encoders (e.g., PCT~\cite{guo2021pct}, Point-BERT~\cite{yu2022point}) further enhanced global context modeling through self-attention, but are mostly trained with geometry-oriented objectives.

More recently, multimodal 3D foundation models such as ULIP~\cite{xue2023ulip}, Point-Bind~\cite{guo2023point}, and PointLLM~\cite{xu2024pointllm} align 3D features with vision and language through contrastive learning, enabling more semantically informed representations.

\subsection{Rationale for Choosing Uni3D as Backbone}
\label{sec:backbone_rationale}
\begin{table}[t]
\centering
\caption{\textbf{Rationale for choosing Uni3D as the semantic backbone.}
Comparison between SGSoft design requirements and the properties supported by Uni3D.}
\label{tab:uni3d_rationale}

\resizebox{\linewidth}{!}{
\begin{tabular}{lcc}
\toprule
\textbf{Design requirement for SGSoft} & \textbf{Required} & \textbf{Supported by Uni3D} \\
\midrule
Object-level semantic priors & \checkmark & \checkmark \\
Point-wise part-level semantics & \checkmark & \checkmark \\
Direct encoding from raw 3D geometry & \checkmark & \checkmark \\
Generalization across object categories and domains & \checkmark & \checkmark \\
\bottomrule
\end{tabular}
}
\end{table}

Among recent multimodal 3D foundation models, \textbf{Uni3D}~\cite{liu2024uni3d} is a large-scale point-cloud foundation model trained with multimodal supervision.  
Although it is not specifically designed for dense surface correspondence, Uni3D provides strong object-level semantic features and structured point-wise representations, as demonstrated by its performance on multiple 3D recognition and segmentation benchmarks.

Moreover, its representations are learned directly from 3D geometry without relying on rendered views, which helps avoid viewpoint-dependent artifacts and preserves local neighborhood relationships in the feature space.  
This spatial consistency is particularly important for correspondence learning, where stable geometric neighborhoods must be preserved under large deformations.

Therefore, we adopt Uni3D as the semantic backbone of SGSoft, given its multimodal pretraining on large-scale 3D data, 
spatially discriminative point-wise feature representations, and stable performance across diverse 3D benchmarks. 
As summarized in Table~\ref{tab:uni3d_rationale}, these properties align well with the design requirements of SGSoft for 
semantically consistent and geometry-aware correspondence learning.

\begin{figure}[t]
  \centering
  \includegraphics[width=\linewidth]{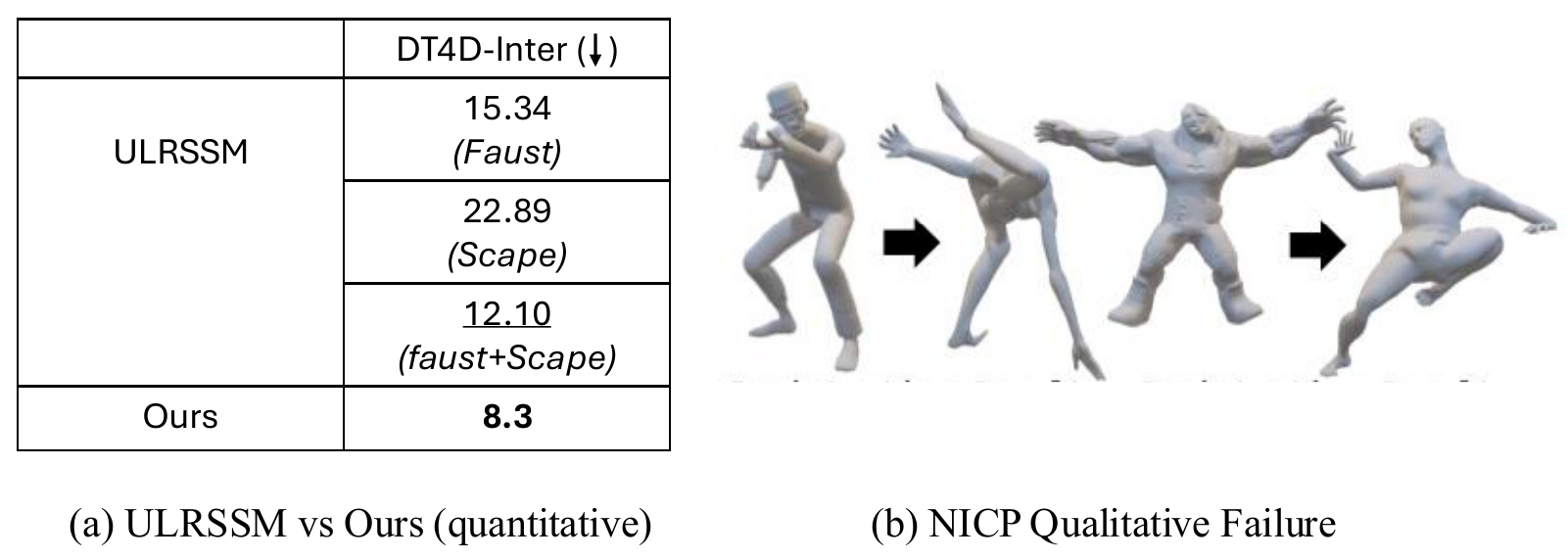}
  \caption{
    \textbf{Comparison with ULRSSM (left) and NICP (right).}
    SGSoft achieves lower error and more consistent correspondences under cross-domain and large deformation settings.
    ULRSSM is not designed for zero-shot correspondence, and NICP follows a different registration objective.
    }

  \label{fig:supp_additionalcomparison}
\end{figure}

\section{Additional Comparison and Uni3D Ablation}
\subsection{Additional Comparison with Other Methods}
We further provide qualitative comparisons with ULRSSM~\cite{cao2023unsupervised} and NICP~\cite{marin2024nicp}, which represent classical unsupervised spectral matching and non-rigid registration approaches, respectively. ULRSSM is an unsupervised spectral matching method, while NICP is designed under relatively strong assumptions on shape similarity and initialization.

As shown in Fig.~\ref{fig:supp_additionalcomparison}, SGSoft produces more stable and semantically consistent correspondences under large deformations and cross-category settings. In contrast, ULRSSM shows limited generalization across domains, and NICP often struggles when the input shapes deviate from its underlying assumptions.

We note that these methods are not specifically designed for zero-shot cross-category correspondence. Nevertheless, we include these comparisons to provide a broader perspective on the behavior of existing approaches under such challenging conditions.

\subsection{Effect of Uni3D Semantic Features}

We further analyze the contribution of semantic features from Uni3D.
Ablating the semantic features from Uni3D (w/o Uni3D) leads to a drastic performance drop across all benchmarks 
(e.g., 2.9$\rightarrow$68.0 on SCAPE), indicating that semantic anchoring is essential for resolving global ambiguities 
and enabling stable correspondence.

\begin{table}[h]
\centering
\caption{Ablation study on Uni3D semantic features.}
\label{tab:ablation_uni3d}
\begin{tabular}{lccc}
\toprule
Method & SCAPE & SHREC19 & DT4D-Inter \\
\midrule
w/o Uni3D & 68.0 & 70.6 & 69.2 \\
Full (SGSoft) & 2.9 & 4.0 & 8.3 \\
\bottomrule
\end{tabular}
\end{table}

\section{Downstream Applications}
\label{sec:applications}

\paragraph{Semantic Segmentation.}
We evaluate semantic label transfer in the SGSoft descriptor space.
As shown in Fig.~1 of the main paper, our descriptors preserve part-level coherence
and disambiguate left-right symmetry, enabling reliable cross-instance label
propagation across different poses and shapes.
These results indicate that SGSoft provides a stable and transferable semantic
representation that can serve as a generic backbone for downstream segmentation tasks.

\begin{figure}[t]
  \centering
  \includegraphics[width=\linewidth]{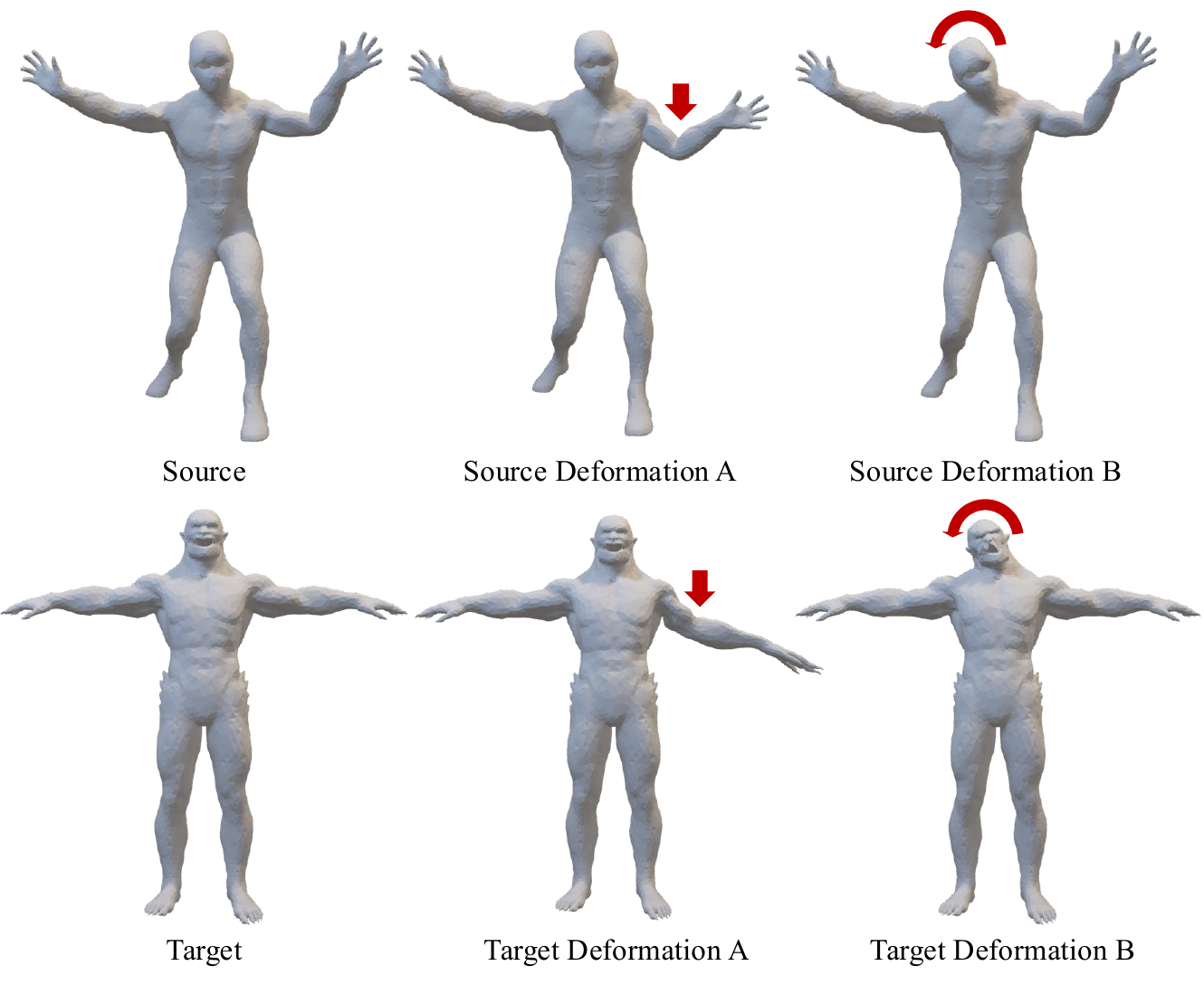}
  \caption{
    \textbf{Deformation transfer with SGSoft correspondences.}
    Given a source mesh and two source deformations (A and B), we propagate the deformations to an unseen target using our predicted correspondences.
    The transferred motions preserve the intended pose changes while adapting smoothly to the target’s body shape.
  }
  \label{fig:supp_deformation}
\end{figure}

\paragraph{Deformation Transfer.}
Using our predicted correspondences, we transfer source deformations to unseen targets. SGSoft yields smoother deformation across joints and avoids typical artifacts such as stretching.  Figure~\ref{fig:supp_deformation} shows two example deformations (A and B) applied to a source and then transferred to a target. The results retain the intended motion while respecting the target’s shape and proportions, indicating that our correspondences are precise enough to drive high-quality deformation transfer.

\section{Limitations and Future Work}
\label{sec:limitation}

\begin{figure}[t] \centering \includegraphics[width=\linewidth]{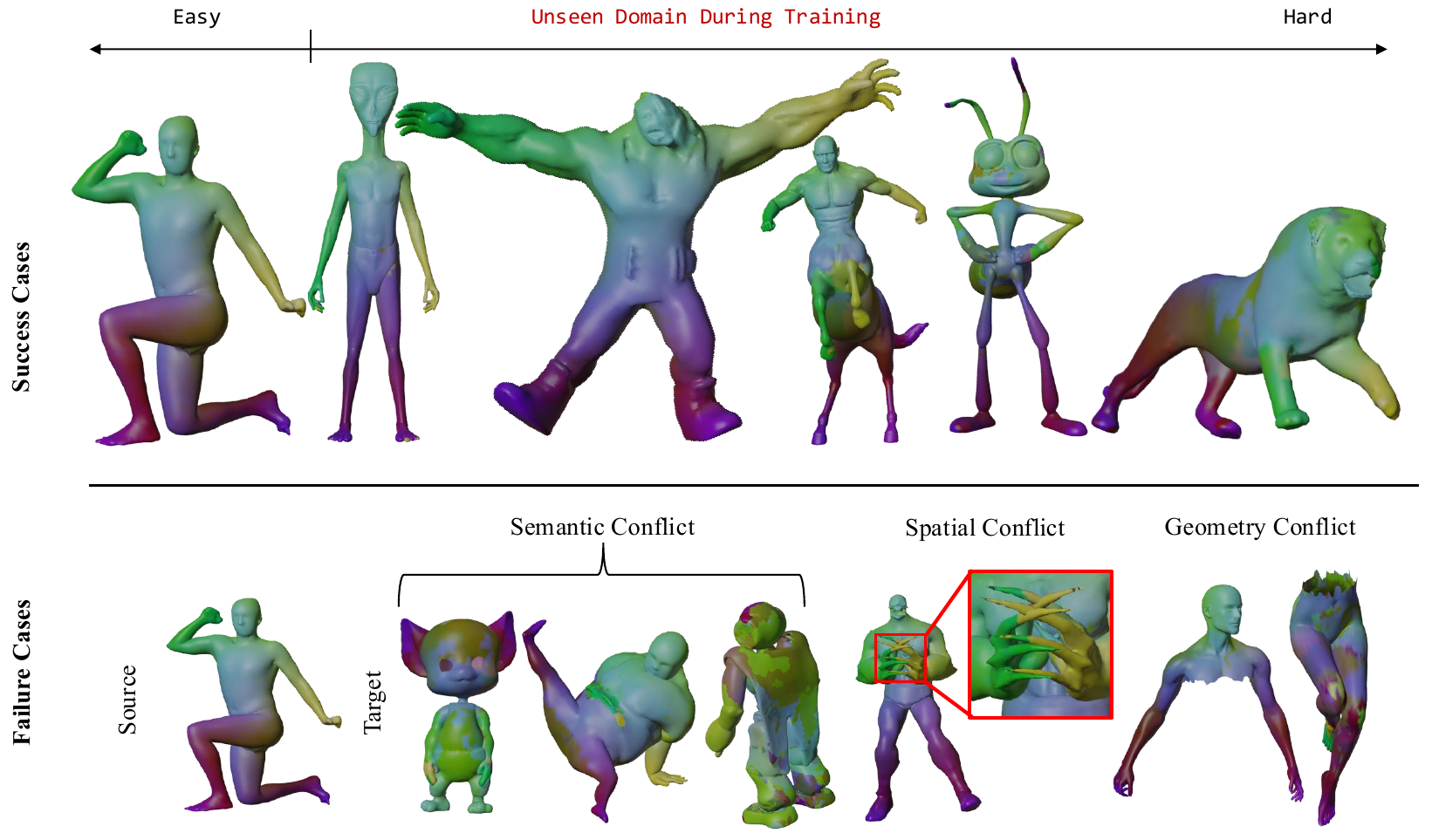} 
\caption{\textbf{Success and failure cases of SGSoft across increasing domain shift.}
Top: Successful correspondences from seen to unseen domains with progressively increasing difficulty.
Bottom: Representative failure modes. (1) Semantic conflict under extreme proportions or additional structures,
(2) spatial ambiguity when parts are in close proximity (zoomed),
(3) geometric degradation on severely partial shapes due to unreliable geodesic distances.}
\label{fig:supp_failure} 
\end{figure}

We identify three representative failure modes, as illustrated in Fig.~\ref{fig:supp_failure}.
First, semantic conflicts arise in shapes with extreme proportions or additional structures (e.g., accessories), where semantic priors become ambiguous.
Second, spatial ambiguity occurs when multiple parts are in close proximity, leading to confusion despite correct global context.
Third, in severely partial shapes, missing connectivity degrades the reliability of geodesic distances, reducing global alignment accuracy.
Overall, SGSoft remains robust under moderate deformation and partiality, with failures primarily occurring when both semantic and geometric cues are unreliable.

While SGSoft demonstrates strong cross-domain generalization, it is still influenced by the choice of canonical template used during training.
In particular, when input shapes exhibit significantly different proportions or structural characteristics from the training domain, the shared reference space induced by the template becomes less optimal.

Extending the framework to support multiple or more diverse templates, or learning a template-agnostic correspondence space, would further improve robustness to large structural variations. We leave this as an important direction for future work.
\begin{figure*}
  \centering
  \includegraphics[width=\textwidth]{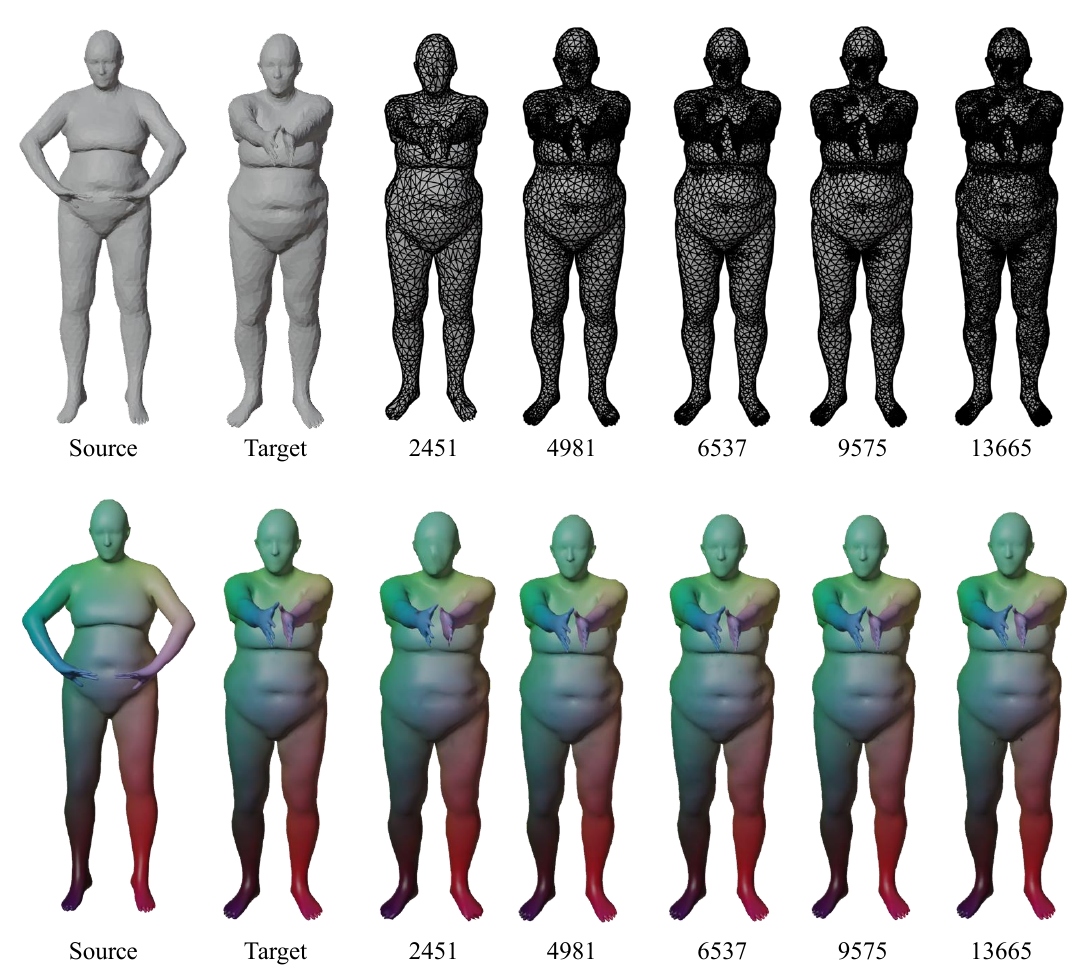}
  \caption{
    \textbf{Topology robustness.}
    Colors are transferred from a source to targets with different mesh resolutions.
    Numbers denote vertex counts.
    SGSoft preserves smooth and semantically consistent correspondences across topologies.
  }
  \label{fig:supp_topology}
\end{figure*}

\begin{figure*}
    \vspace{-6mm}
    \makebox[\textwidth][c]{%
        \hspace{1.5cm} 
        \includegraphics[width=1.5\textwidth]{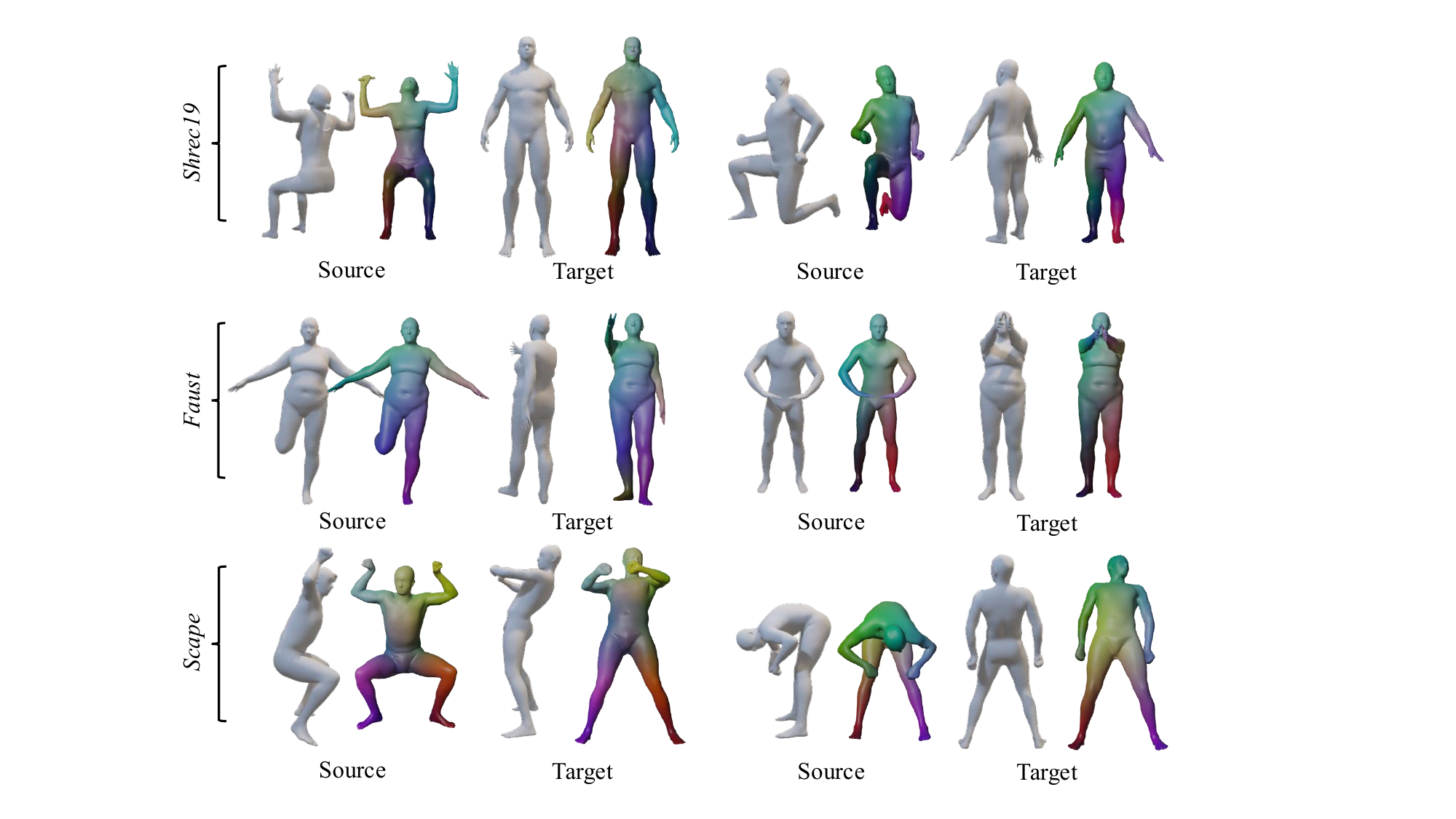}
    }
    \vspace{-5mm}
    \captionof{figure}{
      \textbf{Symmetry disambiguation.}
      We visualize correspondence transfer from a source mesh to target meshes on FAUST, SCAPE, and SHREC'19.
      SGSoft consistently aligns left/right limbs and resolves front-back ambiguity under large articulations, avoiding symmetric flips. 
      The uncolored source and target meshes are shown in their original poses,
      while the colorized meshes are rotated to a front view for clearer visualization of correspondence quality.
    }
    \vspace{-5mm}
    \label{fig:supp_symmetry_disambiguation}
\end{figure*}

\begin{figure*}
    \centering
    \includegraphics[width=0.95\textwidth,height=0.75\textheight,keepaspectratio]{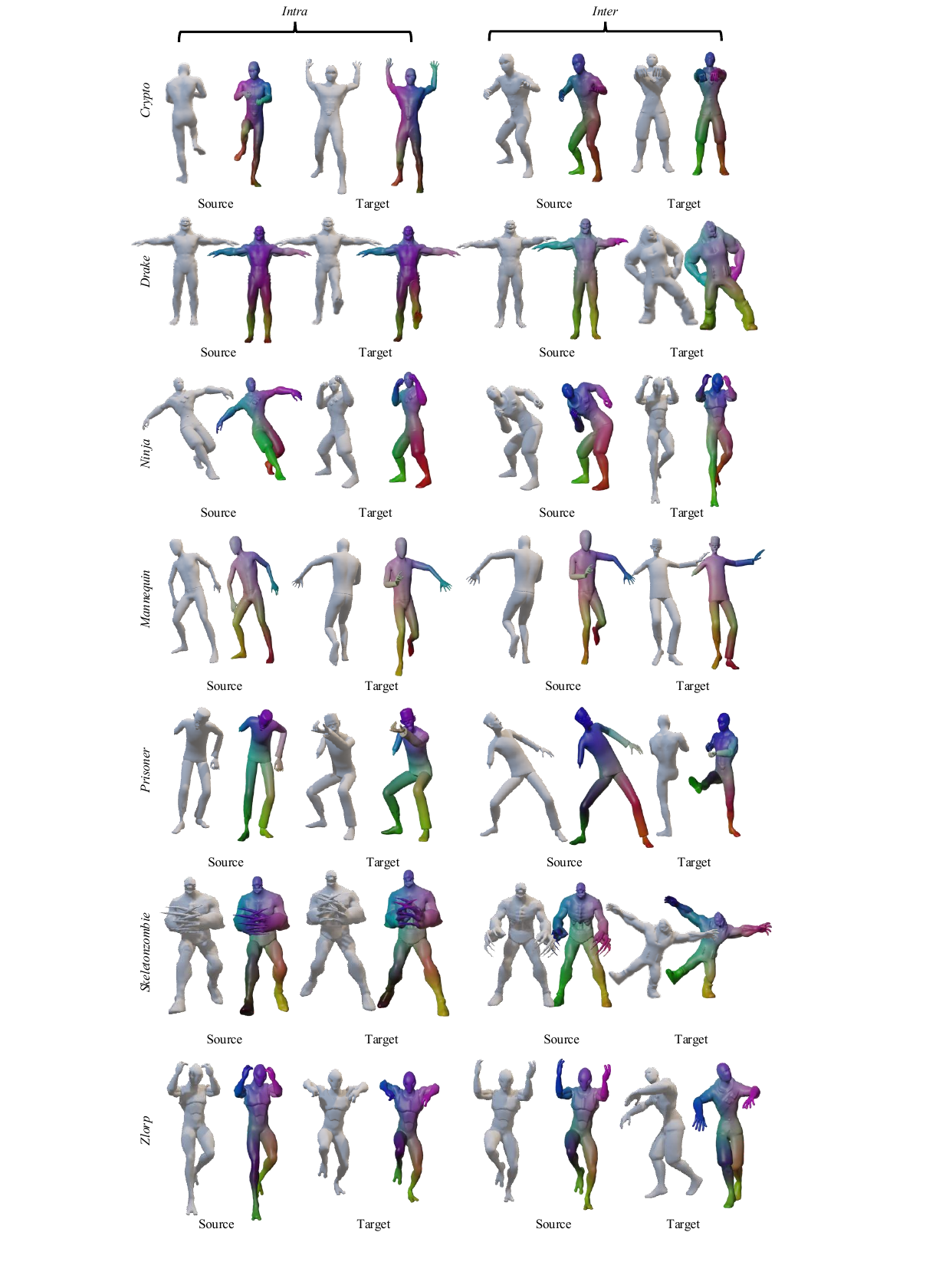}
    \caption{
      \textbf{Zero-shot cross-domain generalization.}
      We train SGSoft only on SMPL-like human bodies and evaluate on stylized rigs from the Mannequin, Drake, Ninja, Prisoner, Skeleton, Zombie, and Crypto Zlorp families.
      \emph{Intra} (left) shows correspondences within each character family, while \emph{Inter} (right) transfers correspondences across different families.
      Despite substantial differences in shape and articulation, SGSoft preserves consistent part-wise color patterns without any finetuning.
    }
    \label{fig:supp_zeroshot}
\end{figure*}


{
    \small
    \bibliographystyle{ieeenat_fullname}
    \bibliography{main}
}

\end{document}